\documentclass[11pt]{article}

\usepackage[preprint]{acl}

\usepackage{times}
\usepackage{latexsym}

\usepackage[T1]{fontenc}

\usepackage[utf8]{inputenc}

\usepackage{microtype}

\usepackage{inconsolata}

\usepackage{graphicx}
\usepackage{booktabs}
\usepackage{multirow}
\usepackage{makecell}
\usepackage{xcolor}
\usepackage[table]{xcolor}
\usepackage{amsmath}
\usepackage{rotating}

\newcommand*{\ourmodel}{C3LM}
\newcommand*{\traindataset}{CREED}

\definecolor{gold}{RGB}{255, 215, 0}
\definecolor{silver}{RGB}{192, 192, 192}
\definecolor{bronze}{RGB}{205, 127, 50}

\title{Training Chemical Plausibility-Aware Large Language Models \\
for Single-Step Retrosynthesis}

\author{
  \textbf{Bogdan Zagribelnyy\textsuperscript{1}},
  \textbf{Ivan Ilin\textsuperscript{1}},
  \textbf{Nikita Bondarev\textsuperscript{1}},
  \textbf{Maksim Kuznetsov\textsuperscript{2}},
\\
  \textbf{Mathieu Reymond\textsuperscript{2}}
  \textbf{Vladimir Aladinskiy\textsuperscript{1}},
  \textbf{Alex Aliper\textsuperscript{1}},
  \textbf{Alex Zhavoronkov\textsuperscript{1,2,3}}
\\
\\
  \textsuperscript{1}Insilico Medicine AI Limited, Abu Dhabi, UAE, \\
  \textsuperscript{2}Insilico Medicine Canada Inc., Montreal, Quebec, Canada, \\
  \textsuperscript{3}Insilico Medicine Hong Kong Ltd., Hong Kong SAR, China \\
\\
  \small{
    \textbf{Correspondence:} \href{mailto:bogdan@insilicomedicine.com}{bogdan@insilicomedicine.com}
  }
}
\begin{document}
\maketitle
\begin{abstract}
Single-step retrosynthesis is a central component of computer-aided synthesis planning, yet its intrinsically one-to-many nature is poorly captured by single-answer evaluation and benchmarking protocols. To address this, we introduce Top-$K$ prompting as a robust training and inference paradigm to better capture diverse, plausible reaction predictions. We compile CREED-CCV-2+USPTO-XL, an ultra-large-scale dataset of $\sim$45.6 million verified reactions to train the C3LM (Chemistry Constraint-Consistent Language Model). By integrating fine-tuning with ChemCensor-based and novelty-oriented rewards, our model achieves state-of-the-art performance on the OOD URSA-expert-2026 benchmark. Further analysis of reaction uniqueness shows that LLMs and conventional models explore complementary reaction spaces, motivating ensemble-based retrosynthesis systems. Overall, our results establish Top-K, plausibility-aware training as a practical new direction for robust future LLM-based synthesis planning.
\end{abstract}

\section{Introduction}
\label{introduction}

\begin{figure}[h]
  \centering
  \includegraphics[
    width=0.90\linewidth,
    trim=15 15 20 0,
    clip
  ]{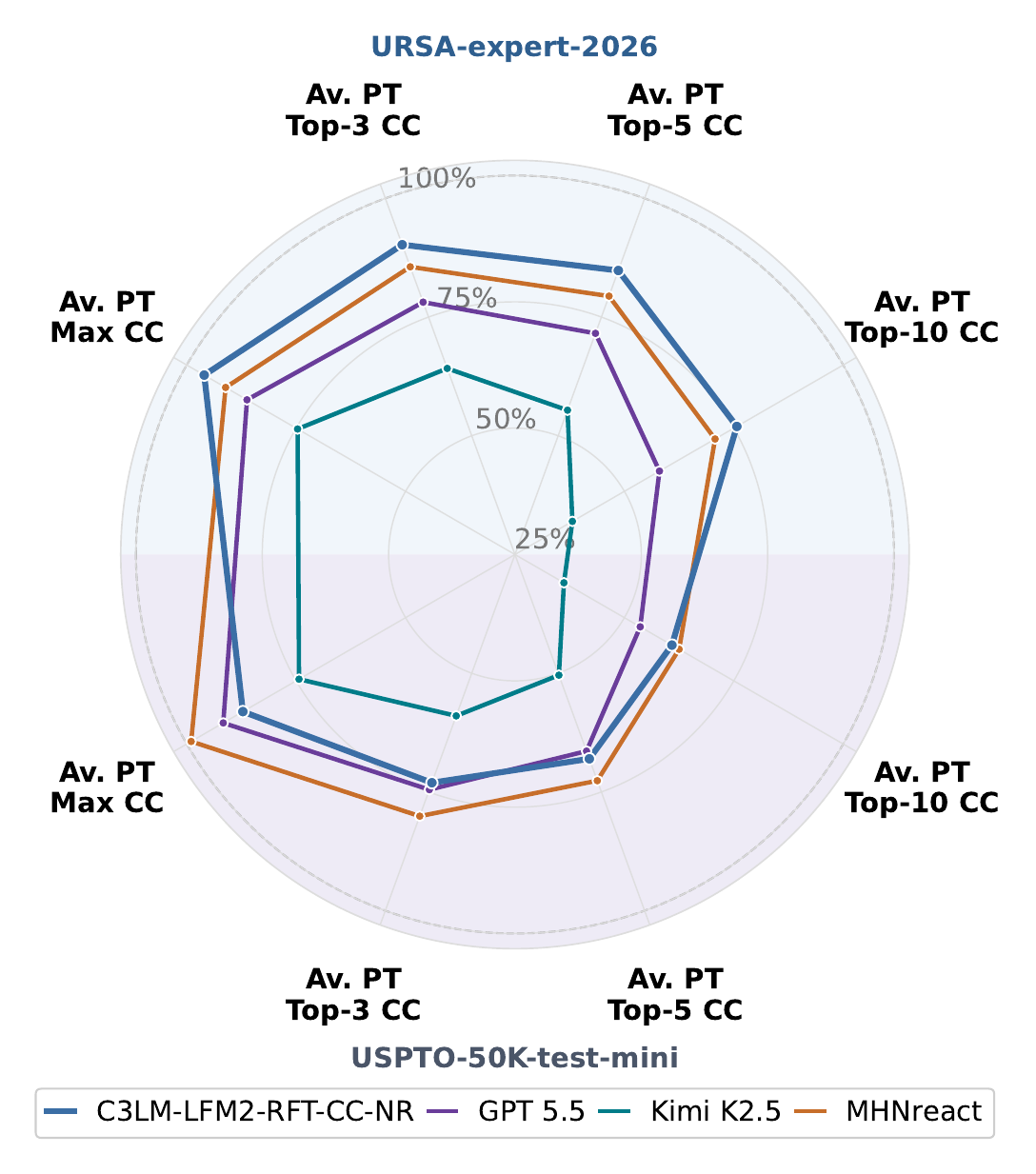}
\caption{Comparison of representative top-performing models from each model tier, normalized by metric-specific frontier scores.}
  \label{fig:radar_frontier_comparison}
\end{figure}

Contemporary small-molecule drug discovery requires careful consideration of whether newly designed compounds are synthetically accessible \citep{gao2020synthesizability}. In real-world medicinal chemistry, the best way to model synthetic accessibility is typically found through synthesis planning \citep{zagribelnyy2026rersa}, with retrosynthetic analysis being the central approach \citep{VLEDUTS1963117, corey1967general}. Computer-aided synthesis planning (CASP) seeks to automate this task by identifying plausible synthetic routes from a target molecule (TM) back to commercially or readily available building blocks \citep{tu2025askcos-buyables}. Although end-to-end multistep retrosynthesis models are beginning to appear \citep{shee2025directmultistep}, many CASP systems still formulate synthesis planning as the problem to be resolved by two connected algorithms: (\textit{i}) a single-step retrosynthesis (SSRS) model, proposing possible reactions and precursors, and (\textit{ii}) a multistep retrosynthesis (MSRS) search engine, where these individual reactions are orchestrated into complete synthetic routes \citep{segler2017ml-in-rs, segler2018-solvability-mcts}. In the past decade, many SSRS models (such as LocalRetro \citep{Chen2021localretro}, MHNreact \citep{Seidl2022mhnreact} and RetroKNN \citep{xie2023retrosknn}) have become conventional tools that can be easily integrated into CASP systems \citep{maziarz2025syntheseus}. At the same time, the rise of Large Language Models (LLMs) \citep{wolf2020transformers, brown2020gpt} and their broad applications to chemistry \citep{Boiko2023-autonomous-llms} in general and synthesis planning in particular \citep{Schwaller2026-synthegy} cannot be ignored. A benchmark between LLMs as SSRS models and conventional tools has been recently performed \citep{zagribelnyy2026chemcensor_c3lm} using the novel URSA framework, which utilizes the ChemCensor metric as the proxy of chemical plausibility, as opposed to the Top-K accuracy metric. This study revealed LLMs as promising SSRS models, but inferior to the best conventional solutions.

In our study, we trained an LLM that is capable of outperforming the best conventional models in the SSRS task on the challenging URSA-expert-2026 benchmark. The novel version of \textbf{C3LM} (\textbf{C}hemistry
\textbf{C}onstraint–\textbf{C}onsistent \textbf{L}anguage \textbf{M}odel) was trained on a newly introduced ultra-large ($\sim$45.6M) CREED-CCV-2+USPTO-XL set of reactions and then fine-tuned using fast ChemCensor-based and novelty-oriented rewards to boost chemical plausibility and diversity of the model's outcomes. 

We deliberately optimize C3LM using ChemCensor-based filtering and rewards. Although this creates partial circularity with evaluation on ChemCensor metrics, the goal is to develop the most practically useful model for end users by directly improving the plausibility and precedent support of generated reactions.

The key contributions of our work are as follows:

{
\begin{enumerate}
    \setlength{\itemsep}{0pt}
    \setlength{\parskip}{0pt}
    \setlength{\parsep}{0pt}
    \setlength{\topsep}{0pt}
    \item We propose Top-$K$ mode for LLM training and prompting as a best practice for increasing the diversity of generated reactions.
    \item We construct CREED-CCV-2+USPTO-XL, a dataset of $\sim 45.6$M verified reactions derived from expert-coded templates.
    \item We train a new version of C3LM on the novel dataset, improving performance over conventional SSRS models and other LLMs.
    \item We perform an in-depth comparison of LLMs and conventional tools from the perspective of reaction uniqueness and identify the current frontiers of the SSRS task.
\end{enumerate}
}

\section{Approach}
\label{sec:approach}

\begin{figure*}[t]
  \centering
  \includegraphics[
    width=\linewidth,
    trim=60 175 70 160,
    clip
  ]{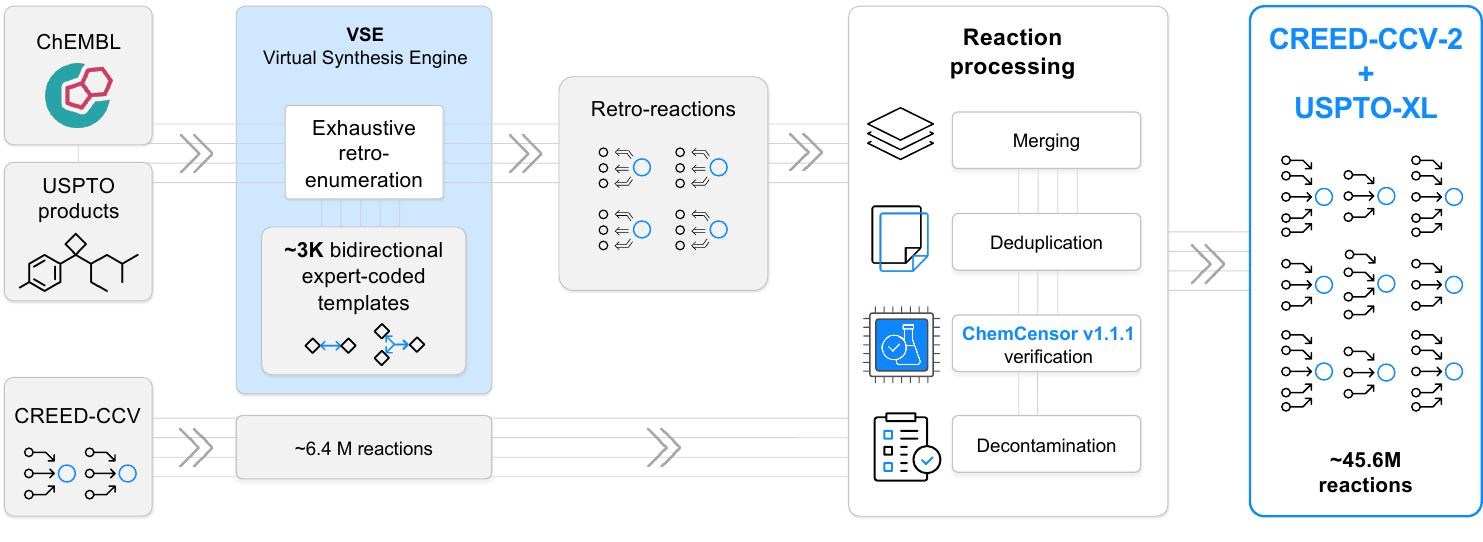}
\caption{Data-generation pipeline for the CREED-CCV-2+USPTO-XL training set.}
  \label{fig:creed-ccv}
\end{figure*}

\subsection{Prompting modes}
To evaluate the capability of Large Language Models (LLMs) in exploring the multi-path nature of single-step retrosynthesis, we formalize the task under two distinct prompting paradigms: Top-1 Prompting Mode and Top-$K$ Prompting Mode. Both modes use a core set of 15 diverse natural language templates adapted from MolInstructions, which wrap the target product's SMILES string.

In the Top-1 Prompting Mode, the model is evaluated under a standard zero-shot configuration. Given a target product SMILES embedded within one of the 15 baseline templates, the LLM is instructed to generate a single, most plausible set of reactants. In one experiment, the model is prompted 15 times by selecting the random template, and all answers are collected, producing 15 reactant sets per target molecule. This mode serves as our baseline. It was used in the previous research of SSRS task on LLMs \citep{zagribelnyy2026chemcensor_c3lm}.

The Top-$K$ Prompting Mode is devoted to simulate behaviour of the conventional SSRS models and capture the inherent one-to-many complexity of retrosynthesis — where a single target molecule can often be potentially synthesized via multiple independent disconnections. To apply this mode, we modify the prompts provided in \citep{zagribelnyy2026chemcensor_c3lm} and append the explicit suffix instruction, "Give me 15 different answers",  to each of the 15 baseline templates from Top-1 Prompting Mode. In one experiment, the LLM is prompted with 3 random templates for each target molecule and up to 15 reactant sets are collected for each template. The predicted outputs are then scored independently and the final metrics are averaged between 3 calculated reaction sets.

\subsection{Training sets}

\textbf{USPTO} is decontaminated USPTO-full \citep{Lowe2017usptofull}, consisting of unique $\sim$897K products and $\sim$951K reactions.

\textbf{CREED-CCV} is generated using a virtual synthesis engine and ChemCensor (v.0.5.2) as a chemical plausibility verification framework and consists of unique $\sim$699K products and $\sim$6.4M reactions \citep{zagribelnyy2026chemcensor_c3lm}.

\textbf{CREED-CCV-2} is collected from two sources: (1) compounds from ChEMBL (v34) \citep{chembl} subjected to the Virtual Synthesis Engine (VSE) to enumerate reactions, (2) reactions from CREED-CCV (see \autoref{fig:creed-ccv}). Both sets of reactions are then merged, deduplicated and scored in ChemCensor (v.1.1.1). Finally, CREED-CCV-2 consists of unique $\sim$2.9M products and $\sim$36M reactions. 

\textbf{USPTO-XL} is generated using unique products from USPTO and the VSE to enumerate reactants in order to address the main limitation of USPTO, that it mainly contains 1 reaction per product. The generated reactants are then verified by ChemCensor (v.1.1.1). The final USPTO-XL set contains unique $\sim$859K products and $\sim$10.6M reactions.

\section{Experiments}
\label{sec:experiments}

\subsection{Baselines}
\label{sec:baselines}

We benchmark a list of proprietary and open-weight foundation general-purpose (GP) LLMs, as well as conventional SSRS models (\autoref{app:baselines}).

\subsection{\ourmodel{} Supervised Fine-Tuning}
\label{sec:sft_train_details}

\textbf{Training setups:}
We perform supervised fine-tuning (SFT) of the \ourmodel{} model under three data configurations. Specifically, we train on either (\textit{i}) CREED-CCV+USPTO, where USPTO was upsampled to match the size of CREED-CCV, or (\textit{ii}) CREED-CCV-2+USPTO-XL that consists of 3{,}680{,}906 unique products and 45{,}649{,}785 unique reactions, which are derived from CREED-CCV-2 and USPTO-XL via merging and  deduplication (\autoref{fig:creed-ccv}). Both models are trained in Top-K mode. The latter C3LM is also asked in its answers to sort the generated reactions according to their predicted plausibility in terms of the ChemCensor score. 

We initialize \ourmodel{}s with \texttt{LFM2 2.6B} checkpoint \citep{amini2025lfm2} and train with basic reasoning for 50{,}000 steps:
\texttt{C3LM-LFM2-CREED-CCV+USPTO} on CREED-CCV+USPTO and
\texttt{C3LM-LFM2-CREED-CCV-2+USPTO-XL} trained on CREED-CCV-2+USPTO-XL, respectively.

The details of the training procedure and training data preprocessing can be found in \autoref{app:c3lm_sft}.

\subsection{\ourmodel{} Reinforcement Learning Fine-Tuning}
\label{sec:rl_train_details}

We perform online Reinforcement Learning Fine-Tuning of the \texttt{C3LM-LFM2-CREED-CCV-2+USPTO-XL} using single-reward Group Relative Policy Optimization (GRPO) \cite{shao2024deepseekmathpushinglimitsmathematical}. RFT uses the same \traindataset{} training split as SFT with a GRPO group size of 8, sampling temperature of $1$, and KL-regularization weight of 0.1. The GRPO reward function is defined as a weighted sum combination of 6 different reward components. Firstly, 3 components are used for syntax purposes, i.e., thinking format validity, generation of valid SMILES strings, and generation of exactly $k$ answers. Next, to avoid duplicates, 1 component encourages the generation of unique solutions. Finally, the main components aim to generate chemically plausible and novel reactants. For plausibility, we maximize the ChemCensor score of each generated reactant. For novelty, we encourage the generation of reactants outside of the exhaustive lists of reactants of CREED-CCV-2+USPTO-XL that also exhibit a positive ChemCensor score. Details about training and each of the reward components are available in \autoref{app:c3lm_rft}. 
The inventory of both SFT and RFT C3LM models can be found in \autoref{app:inventory}.

\subsection{Evaluation protocol}

\begin{figure}[t]
  \centering
  \includegraphics[
    width=\linewidth,
    trim=0 0 0 0,
    clip
  ]{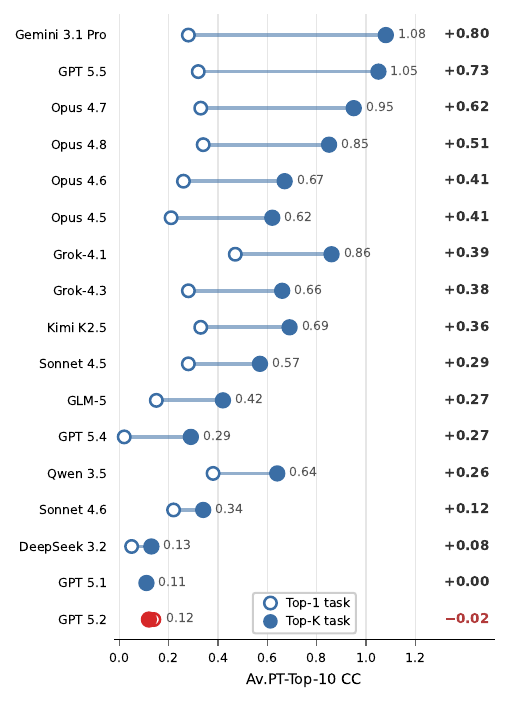}
\caption{Top-1 $\to$ Top-K transition on URSA-expert-2026 (Av.\ PT-Top-10 CC). Hollow = Top-1 task, filled = Top-K task; line length is the gain $\Delta = \text{Top-K} - \text{Top-1}$ (right column, sorted).}
  \label{fig:topk_top1_comparison_}
\end{figure}

We perform a benchmark of all models from Sec. \ref{sec:baselines} and \ourmodel{}s on the recently proposed URSA-expert-2026 and USPTO-50K-test-mini sets \citep{zagribelnyy2026chemcensor_c3lm} using ChemCensor (v.1.1.1). In case of LLMs, for each product, we generate either (\textit{i}) 15 independent responses using 15 prompts (Top-1 prompting mode) or (\textit{ii}) 3 independent responses from each model using 1 random Top-15 prompt (1 random prompt to get 15 reactions, i.e. Top-$K$ prompting mode, $K$=15). Then we average CC-aggregated metrics \textbf{Av. PT-Max CC} and \textbf{Av. PT-Top-K} reported in \citep{zagribelnyy2026chemcensor_c3lm} by 3. In case of conventional SSRS models, 15 reactions are generated per product using the models installed via Syntheseus \citep{maziarz2025syntheseus}. Then the reactions are evaluated using CC-aggregated metrics. The benchmark was carried out using two databases of synthetic precedents: major (\textit{i}) public USPTO-full and supplemental (\textit{ii}) combined USPTO-full with commercial Pistachio set \citep{pistachio_nextmove} (see \autoref{app:u2p2}).

To study different frontiers of reactions' plausibility and diversity, all answers from all respective models are collected together with CC-based sorting applied to the obtained list of reactions for each TM. After that, the best 15 reactant sets are retained and CC-metrics are calculated as usual.
\section{Results}

\begin{table*}[th!]
\centering
\begin{tabular}{@{}l|cccc|cccc@{}}
\toprule
\multirow{3}{*}{\textbf{Model}} &
\multicolumn{4}{c|}{\textbf{URSA-expert-2026}} &
\multicolumn{4}{c}{\textbf{USPTO-50K-test-mini}} \\
\cline{2-5}\cline{6-9}
& \multirow{2}{*}{\textbf{Max}} & \multicolumn{3}{c|}{\textbf{Av. PT-Top-K CC}}
& \multirow{2}{*}{\textbf{Max}} & \multicolumn{3}{c}{\textbf{Av. PT-Top-K CC}} \\
&  & \textbf{@3} & \textbf{@5} & \textbf{@10}
&  & \textbf{@3} & \textbf{@5} & \textbf{@10} \\
\midrule
\multicolumn{9}{c}{\textit{Proprietary Foundation Models}} \\
\midrule
Grok-4.1                   & 1.86 & 1.56 & 1.31 & 0.86 & 3.99 & 2.61 & 2.02 & 1.24\\
Grok-4.3                   & 1.74 & 1.41 & 1.13 & 0.66 & 3.99 & 2.58 & 1.95 & 1.15\\
Gemini 3.1 Pro             & 1.91 & 1.69 & 1.46 & 1.08 & 4.32 & 2.92 & 2.34 & 1.59 \\
GPT 5.4                    & 1.19 & 0.78 & 0.55 & 0.29 & 2.34 & 1.43 & 1.02 & 0.55 \\
GPT 5.5                    & 1.94 & 1.68 & 1.46 & 1.05 & 4.50 & 3.07 & 2.45 & 1.63 \\
Claude Opus 4.7            & 1.91 & 1.65 & 1.39 & 0.95 & 4.35 & 2.96 & 2.31 & 1.45 \\
Claude Opus 4.8            & 1.89 & 1.62 & 1.34 & 0.85 & 4.35 & 2.95 & 2.30 & 1.44 \\

\midrule
\multicolumn{9}{c}{\textit{Open-weight Foundation Models}} \\
\midrule
Qwen 3.5                   & 1.61 & 1.32 & 1.07 & 0.64 & 3.44 & 2.39 & 1.84 & 1.09\\
Kimi K2.5                  & 1.68 & 1.38 & 1.13 & 0.69 & 3.65 & 2.43 & 1.86 & 1.10\\
GLM-5                      & 1.22 & 0.97 & 0.75 & 0.42 & 2.16 & 1.39 & 1.03 & 0.58\\
\midrule
\multicolumn{9}{c}{\textit{Conventional SSRS Models}} \\
\midrule
LocalRetro          & 2.11 & \cellcolor{bronze!40}1.85 & 1.59 & 1.22 & 4.84 & \cellcolor{bronze!40}3.31 & \cellcolor{silver!40}2.67 & 1.81 \\
GLN                 & 1.96 & 1.72 & 1.49 & 1.03 & 4.80 & 3.18 & \cellcolor{bronze!40}2.52 & 1.58 \\
MHNreact            & 2.05 & 1.84 & \cellcolor{bronze!40}1.62 & \cellcolor{bronze!40}1.28 & 4.86 & 3.30 & \cellcolor{gold!40}2.68 & \cellcolor{gold!40}1.90 \\
RetroKNN            & 2.10 & 1.84 & 1.60 & 1.22 & 4.85 & \cellcolor{silver!40}3.33 & \cellcolor{gold!40}2.68 & \cellcolor{bronze!40}1.82 \\
R-SMILES            & 2.08 & 1.83 & 1.56 & 1.11 & 4.85 & \cellcolor{gold!40}3.36 & \cellcolor{silver!40}2.67 & 1.75 \\
\midrule
\multicolumn{9}{c}{\textit{\ourmodel{}}, Supervised Fine-Tuning, \textbf{Top-1 Mode}} \\
\midrule
\ourmodel{}-LFM2-CREED-CCV+USPTO\textsuperscript{*}              & 1.62 & 1.06 & 0.72 & 0.38 & 4.12 & 2.10 & 1.36 & 0.70 \\
\midrule
\multicolumn{9}{c}{\textit{\ourmodel{}}, Supervised and Reinforcement Learning Fine-Tuning, \textbf{Top-K Mode}} \\
\midrule
\ourmodel{}-LFM2-CREED-CCV+USPTO         & 1.92 & 1.68 & 1.42 & 0.98 & 4.16 & 2.74 & 2.17 & 1.42 \\
\ourmodel{}-LFM2-CREED-CCV-2+USPTO-XL    & 2.04 & 1.81 & 1.59 & 1.27 & 4.16 & 2.88 & 2.37 & 1.70 \\ 
\ourmodel{}-LFM2-RFT-CC          & 2.08 & \cellcolor{silver!40}1.88 & \cellcolor{silver!40}1.65 & \cellcolor{silver!40}1.29 & 4.16 & 2.92 & 2.41 & 1.73 \\
\ourmodel{}-LFM2-RFT-CC-NR       & 2.16 & \cellcolor{gold!40}1.94 & \cellcolor{gold!40}1.73 & \cellcolor{gold!40}1.37 & 4.28 & 3.01 & 2.51 & \cellcolor{silver!40}1.85 \\
\midrule
\multicolumn{9}{c}{\textit{Chemical Plausibility and Diversity Frontier of Generated Reactions}} \\
\midrule
All GP LLMs (17) together                        & 2.19 & 2.07 & 1.93 & 1.65 & 4.85   & 3.98   & 3.53   & 2.87   \\
All conventional SSRS models (8) together             & 2.18 & 1.99 & 1.78 & 1.45 & 4.90   & 3.58   & 3.00   & 2.26   \\
All C3LMs (5) together              & 2.23   & 2.04   & 1.88   & 1.57   & 4.74   & 3.31   & 2.78   & 2.14   \\
All benchmarked models (30) together              & 2.25   & 2.15   & 2.04   & 1.81   & 4.91   & 4.12   & 3.69   & 3.04   \\
\bottomrule
\end{tabular}

\caption{Plausibility-based evaluation in the single-step retrosynthesis Top-$K$ mode. \textbf{Max}: per-target maximum ChemCensor score averaged over TMs. \textbf{Av.\ PT-Top-K CC}: per-TM average ChemCensor score over top-K unique predictions; ChemCensor v1.1.1, USPTO-full as the source of synthetic precedents. See the full table in \autoref{app:full_table}. Results for recently released models are available at \url{https://dddbench.insilico.com/}.}
\label{tab:main_table_topk}
\end{table*}

\textbf{Top-K vs Top-1 prompting:} Benchmarking foundation models reveals broad differentiation between the LLMs in how they are influenced by the prompting framework. Overall results for Top-$K$ mode are presented in \autoref{tab:main_table_topk}, for Top-1 in \autoref{tab:top1} of \autoref{app:single_anwer} respectively. Across the CC-aggregated metrics, \textbf{Av. PT-Top-10} metric is the most sensitive to the diversity of generated reactions, thus the transition from Top-1 to the Top-$K$ mode should be the most descriptive in the case of this metric. As we can see in \autoref{fig:topk_top1_comparison_}, the vast majority of models benefit significantly from this transition. Moreover, the models' rankings change: Grok-4.1 produces the most plausible and diverse reactions in Top-1 mode, while Gemini 3.1 Pro's performance improves dramatically in Top-15 mode, making it the best LLM baseline. Only GPT 5.2 performs worse in the Top-$K$ mode according to \textbf{Av. PT-Top-10} values, while for other CC-based metrics, GPT 5.1 and Claude Sonnet 4.6 also show inferior performance in the Top-$K$ mode (see \autoref{tab:main_table_topk_full} and \autoref{tab:top1}). Finally, the substantial improvements in metrics due to the change in prompting mode force us to establish the best practice for LLM benchmarking in the SSRS task: use the Top-$K$ mode instead of the Top-1 mode. 

\textbf{Top-K mode for the C3LM training:} Considering the Top-$K$ mode is beneficial not only for LLMs' inference, but also for model training, we have trained a \texttt{C3LM-LFM2-CREED-CCV+USPTO} in the Top-$K$ mode and compared it to the model trained on the same dataset but in the Top-1 mode. We can see (\autoref{tab:main_table_topk}) that all CC-based metrics are boosted due to the training mode transition to the Top-$K$ mode, and the most substantial improvement (> 2.5 fold) is observed for \textbf{Av. PT-Top-10} metric, pointing to the increased diversity of the generated reactions. Specifically, the matched Top-1 $\to$ Top-$K$ transition improves Max/@3/@5/@10 by $+0.30/+0.62/+0.70/+0.60$, respectively. As a result of this comparison, the new C3LMs are trained in the Top-$K$ mode. This stage allowed the new SFT C3LM to reach the performance level of top-tier proprietary GP LLMs: Gemini 3.1 Pro and GPT 5.5. 

\textbf{Scaling training set and RFT:} The next significant improvement in C3LMs training is achieved when the training set size increased by > 6 times, when transited from CREED-CCV+USPTO to CREED-CCV-2+USPTO-XL. This dataset transition further improves Max/@3/@5/@10 by $+0.12/+0.13/+0.17/+0.29$, respectively. The resulting \texttt{C3LM-LFM2-CREED-CCV-2+USPTO-XL} is then fine-tuned in RL mode using the ChemCensor (CC) and Novelty (NR) rewards. The use of the CC-reward allows to beat conventional SSRS models on the OOD URSA-expert-2026 benchmark according to \textbf{Av. PT-Top-K CC} metrics, while adding the NR results in the RFT-model dominance across all CC metrics on this benchmark set. CC-RFT adds $+0.04/+0.07/+0.06/+0.02$ to Max/@3/@5/@10, while the novelty reward contributes a further $+0.08/+0.06/+0.08/+0.08$, respectively. From the perspective of the representative sample from the conventional USPTO-50K-test benchmark (USPTO-50K-test-mini), the \textbf{Av. PT-Top-10} metric is the least vulnerable to data leakage, since the vast majority (481/497) of product molecules from the USPTO-50K-test-mini have no more than $3$ reactions in the public USPTO-full (see \autoref{app:dist_average}). In these conditions, the second-best result of \texttt{C3LM-LFM2-RFT-CC-NR} on the USPTO-50K-test-mini benchmark (\textbf{Av. PT-Top-10}) is the additional sign of the models' capabilities to generate genuinely chemically plausible reactions rather than to memorize right answers. Oppositely, the \textbf{Av. PT-Max CC} values close to $5$ (i.e., exact match to the synthetic precedent) by the conventional SSRS models reveal potential leakage of USPTO-50K-based benchmarks to these models and this particular metric on USPTO-50K-test-mini benchmark should be cautiously considered. 

\begin{figure}[t]
  \centering
  \includegraphics[
    width=\linewidth,
    trim=0 10 0 0,
    clip
  ]{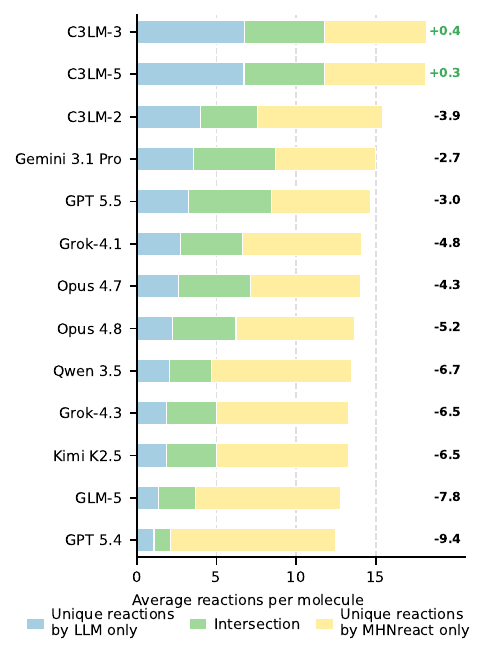}
\caption{Intersection between the reactions predicted for the URSA-expert-2026 benchmark set by each LLM and those predicted by MHNreact. Values on the right ($\Delta$) show the average per-target difference between the number of reactions unique to the LLM and those unique to MHNreact; positive values indicate more LLM-unique reactions. \textbf{C3LM-2}: \texttt{C3LM-LFM2-CREED-CCV+USPTO}; \textbf{C3LM-3}: \texttt{C3LM-LFM2-CREED-CCV-2+USPTO-XL}; \textbf{C3LM-5}: \texttt{C3LM-LFM2-RFT-CC-NR}. See all results in \autoref{fig:intersection_full}.}
  \label{fig:intersection}
\end{figure}

\textbf{Frontier of plausibility and diversity}: While \texttt{C3LM-LFM2-RFT-CC-NR} achieves results on the challenging benchmark, the real frontier of chemical plausibility and diversity of generated reactions remains unknown due to the one-to-many nature of the ChemCensor-based benchmarks. Understanding the frontier values would be helpful for the next iterations of the research to navigate the efforts. Collecting the best answers across GP LLMs reveals that they together produce more plausible and diverse outcomes than the conventional SSRS models, while the best individual GP models (Gemini 3.1 Pro and GPT 5.5) are still far from the best conventional models (LocalRetro and MHNreact). The collected and sorted answers from all C3LMs also allow for higher values of CC-aggregated metrics than for conventional models, while all GP LLMs together are slightly superior to all C3LMs together. The ultimate detectable frontier values of the CC-aggregated metrics that can be identified so far have been calculated from the reactions generated by all 30 benchmarked models (see \autoref{tab:main_table_topk}). These values show that there is still room for improvement in models, especially with respect to metrics that are sensitive to both reactions' diversity and plausibility (+0.44 for \textbf{Av. PT-Top-10} and +0.31 for \textbf{Av. PT-Top-5}). Finally, the \% of frontier values of the CC-based metrics achieved by a model can be considered as an additional metric to visualize the performance of the model (see \autoref{fig:radar_frontier_comparison}). 

\textbf{The analysis of reactions}: Looking at the values of CC-aggregated metrics from the \autoref{tab:main_table_topk}, the genuine chemical contributions in terms of proposing unique plausible reactions might remain unclear. In order to reveal the capabilities of LLMs to provide unique reactions at the top of the performance by SSRS models, we compare reactions generated by the MHNreact model, which is the most capable of generating diverse outputs (best \textbf{Av. PT-Top-10} values across conventional models), to the reactions generated by the LLMs on the URSA-expert-2026 benchmark. The full results of models' comparison from the reactions' uniqueness perspective for URSA-expert-2026 are available in \autoref{app:model_intersection}, while key highlights can be observed in \autoref{fig:intersection}. MHNreact dominates across the vast majority of LLMs to give plausible reactions (ChemCensor Score > 0) that have not been generated by them. Only models that produce more unique reactions than MHNreact are \texttt{C3LM-LFM2-CREED-CCV-2+USPTO-XL} (+0.4 reactions on average) and \texttt{C3LM-LFM2-RFT-CC-NR} (+0.3 reactions on average). The examples of reactions generated by \texttt{C3LM-LFM2-CREED-CCV-2+USPTO-XL} and MHNreact for one of the target molecules from the benchmark set can be found in \autoref{fig:intersection_example} of \autoref{app:model_intersection}. At the same time, we can see that every LLM can predict some plausible reactions that have not been proposed by MHNreact, and the proportion of such reactions, as well as the proportion of intersecting reactions, generally increases with newer versions of GP LLMs. These findings highlight that the chemical spaces of plausible reactions generated by the best conventional SSRS model and LLMs are quite different, and this can motivate researchers to improve both LLMs and conventional models and/or to use the ensemble of models (conventional model and LLM) to cover a broader chemical space of generated reactions.

\section{Ethical considerations}

The development and deployment of computer-aided synthesis planning (CASP) tools carry significant dual-use potential, as technologies capable of automating the design of synthetic routes for complex small molecules could, in principle, be repurposed to facilitate the synthesis of hazardous compounds. Our research exemplifies the usage of the CREED dataset and the ChemCensor scoring in the pursuit of improving the objective evaluation of chemical plausibility and synthetic feasibility for drug discovery. While our work improves the reliability and interpretability of CASP tools and their more robust benchmarking methodology, we recognize that the ability to accurately plan synthetic routes must be coupled with responsible stewardship.

To mitigate potential risks, we have focused our benchmarking efforts on the legitimate goal of automating synthesis planning for medicinal chemistry, specifically utilizing novel molecular structures that do not overlap with known hazardous compound databases. Furthermore, the ChemCensor framework is built upon established open-scientific data from patent literature and does not inherently possess or provide instructions for the synthesis of toxic or regulated substances. We emphasize that all applications of LLM-based retrosynthesis should be conducted within established institutional biosafety frameworks and comply with relevant international and local regulations governing chemical research. We advocate for the ongoing development of safety-aligned AI systems that incorporate robust safeguards to prevent the generation of unauthorized synthetic pathways, ensuring that the advancement of AI in drug discovery remains focused on beneficial scientific and clinical outcomes.

\section{Limitations}

Despite the performance gains achieved by our Top-$K$, plausibility-aware training paradigm, this study is subject to several limitations that warrant future investigation. As already mentioned in \autoref{introduction}, we acknowledge a degree of alignment between our optimization criteria and our evaluation metrics. Nevertheless, we contend that this "circularity" is a functional design choice: our primary aim is to maximize the model's utility for practitioners. While the ChemCensor serves as an efficient proxy for chemical plausibility, it still does not fully account for practical laboratory organic synthesis parameters, such as reaction conditions, solvents, and purification methods. The reference datasets of synthetic precedents used for ChemCensor scoring in the frame of this study are limited to patent-derived reaction spaces. Chemical diversity is assessed solely through exact SMILES string matching, which lacks a chemically grounded comparison of the predicted transformations, such as reaction class or mechanistic type. Finally, we understand that the generation of our training dataset relies on a template-based engine, which might be biased toward particular chemical patterns and may lack some omitted and "novel-chemistry" transformations.
\section{Code and Benchmark Availability}
\label{sec:availability}

The ChemCensor source code is publicly available at
\url{https://github.com/insilicomedicine/ChemCensor}.

Updated benchmarking results, including evaluations of recently released models, are available at
\url{https://dddbench.insilico.com/}.

\newpage

\bibliography{bibliography}

@Article{livne2024nacho,
author ="Livne, Micha and Miftahutdinov, Zulfat and Tutubalina, Elena and Kuznetsov, Maksim and Polykovskiy, Daniil and Brundyn, Annika and Jhunjhunwala, Aastha and Costa, Anthony and Aliper, Alex and Aspuru-Guzik, Alán and Zhavoronkov, Alex",
title  ="nach0: multimodal natural and chemical languages foundation model",
journal  ="Chem. Sci.",
year  ="2024",
volume  ="15",
issue  ="22",
pages  ="8380-8389",
publisher  ="The Royal Society of Chemistry",
doi  ="10.1039/D4SC00966E",
url  ="http://dx.doi.org/10.1039/D4SC00966E"}

@inproceedings{brown2020gpt,
 author = {Brown, Tom and Mann, Benjamin and Ryder, Nick and Subbiah, Melanie and Kaplan, Jared D and Dhariwal, Prafulla and Neelakantan, Arvind and Shyam, Pranav and Sastry, Girish and Askell, Amanda and Agarwal, Sandhini and Herbert-Voss, Ariel and Krueger, Gretchen and Henighan, Tom and Child, Rewon and Ramesh, Aditya and Ziegler, Daniel and Wu, Jeffrey and Winter, Clemens and Hesse, Chris and Chen, Mark and Sigler, Eric and Litwin, Mateusz and Gray, Scott and Chess, Benjamin and Clark, Jack and Berner, Christopher and McCandlish, Sam and Radford, Alec and Sutskever, Ilya and Amodei, Dario},
 booktitle = {Advances in Neural Information Processing Systems},
 editor = {H. Larochelle and M. Ranzato and R. Hadsell and M.F. Balcan and H. Lin},
 pages = {1877--1901},
 publisher = {Curran Associates, Inc.},
 title = {Language Models are Few-Shot Learners},
 url = {https://proceedings.neurips.cc/paper_files/paper/2020/file/1457c0d6bfcb4967418bfb8ac142f64a-Paper.pdf},
 volume = {33},
 year = {2020}
}

@article{irwin2022chemformer,
doi = {10.1088/2632-2153/ac3ffb},
url = {https://dx.doi.org/10.1088/2632-2153/ac3ffb},
year = {2022},
month = {jan},
publisher = {IOP Publishing},
volume = {3},
number = {1},
pages = {015022},
author = {Ross Irwin and Spyridon Dimitriadis and Jiazhen He and Esben Jannik Bjerrum},
title = {{C}hemformer: a pre-trained transformer for computational chemistry},
journal = {Machine Learning: Science and Technology}
}

@article{degen2008brics,
author = {Degen, Jörg and Wegscheid-Gerlach, Christof and Zaliani, Andrea and Rarey, Matthias},
title = {On the Art of Compiling and Using 'Drug-Like' Chemical Fragment Spaces},
journal = {ChemMedChem},
volume = {3},
number = {10},
pages = {1503-1507},
doi = {https://doi.org/10.1002/cmdc.200800178},
url = {https://chemistry-europe.onlinelibrary.wiley.com/doi/abs/10.1002/cmdc.200800178},
eprint = {https://chemistry-europe.onlinelibrary.wiley.com/doi/pdf/10.1002/cmdc.200800178},
year = {2008}
}

@inproceedings{wolf2020transformers,
    title = "{T}ransformers: State-of-the-Art Natural Language Processing",
    author = "Wolf, Thomas  and
      Debut, Lysandre  and
      Sanh, Victor  and
      Chaumond, Julien  and
      Delangue, Clement  and
      Moi, Anthony  and
      Cistac, Pierric  and
      Rault, Tim  and
      Louf, Remi  and
      Funtowicz, Morgan  and
      Davison, Joe  and
      Shleifer, Sam  and
      von Platen, Patrick  and
      Ma, Clara  and
      Jernite, Yacine  and
      Plu, Julien  and
      Xu, Canwen  and
      Le Scao, Teven  and
      Gugger, Sylvain  and
      Drame, Mariama  and
      Lhoest, Quentin  and
      Rush, Alexander",
    editor = "Liu, Qun  and
      Schlangen, David",
    booktitle = "Proceedings of the 2020 Conference on Empirical Methods in Natural Language Processing: System Demonstrations",
    month = oct,
    year = "2020",
    address = "Online",
    publisher = "Association for Computational Linguistics",
    url = "https://aclanthology.org/2020.emnlp-demos.6",
    doi = "10.18653/v1/2020.emnlp-demos.6",
    pages = "38--45",
}

@article{weininger1988smiles,
author = {Weininger, David},
title = {{SMILES}, a chemical language and information system. 1. Introduction to methodology and encoding rules},
journal = {Journal of Chemical Information and Computer Sciences},
volume = {28},
number = {1},
pages = {31-36},
year = {1988},
doi = {10.1021/ci00057a005},
URL = {https://doi.org/10.1021/ci00057a005},
eprint = {https://doi.org/10.1021/ci00057a005}
}

@inproceedings{pei2023biot5,
    title = "{B}io{T}5: Enriching Cross-modal Integration in Biology with Chemical Knowledge and Natural Language Associations",
    author = "Pei, Qizhi  and
      Zhang, Wei  and
      Zhu, Jinhua  and
      Wu, Kehan  and
      Gao, Kaiyuan  and
      Wu, Lijun  and
      Xia, Yingce  and
      Yan, Rui",
    editor = "Bouamor, Houda  and
      Pino, Juan  and
      Bali, Kalika",
    booktitle = "Proceedings of the 2023 Conference on Empirical Methods in Natural Language Processing",
    month = dec,
    year = "2023",
    address = "Singapore",
    publisher = "Association for Computational Linguistics",
    url = "https://aclanthology.org/2023.emnlp-main.70",
    doi = "10.18653/v1/2023.emnlp-main.70",
    pages = "1102--1123",
}

@article{corey1967general,
url = {https://www.degruyterbrill.com/document/doi/10.1351/pac196714010019/html},
title = {General methods for the construction of complex molecules},
author = {E. J. Corey},
pages = {19--38},
volume = {14},
number = {1},
journal = {Pure and Applied Chemistry},
doi = {doi:10.1351/pac196714010019},
year = {1967},
lastchecked = {2026-01-22}
}

@article{gao2020synthesizability,
author = {Gao, Wenhao and Coley, Connor W.},
title = {The Synthesizability of Molecules Proposed by Generative Models},
journal = {Journal of Chemical Information and Modeling},
volume = {60},
number = {12},
pages = {5714-5723},
year = {2020},
doi = {10.1021/acs.jcim.0c00174},
URL = {https://doi.org/10.1021/acs.jcim.0c00174},
eprint = {https://doi.org/10.1021/acs.jcim.0c00174}
}

@article{shee2025directmultistep,
author = {Shee, Yu and Morgunov, Anton and Li, Haote and Batista, Victor S.},
title = {DirectMultiStep: Direct Route Generation for Multistep Retrosynthesis},
journal = {Journal of Chemical Information and Modeling},
volume = {65},
number = {8},
pages = {3903-3914},
year = {2025},
doi = {10.1021/acs.jcim.4c01982},
URL = {https://doi.org/10.1021/acs.jcim.4c01982},
eprint = {https://doi.org/10.1021/acs.jcim.4c01982}
}

@Article{maziarz2025syntheseus,
author ="Maziarz, Krzysztof and Tripp, Austin and Liu, Guoqing and Stanley, Megan and Xie, Shufang and Gaiński, Piotr and Seidl, Philipp and Segler, Marwin H. S.",
title  ="Re-evaluating retrosynthesis algorithms with Syntheseus",
journal  ="Faraday Discuss.",
year  ="2025",
volume  ="256",
issue  ="0",
pages  ="568-586",
publisher  ="The Royal Society of Chemistry",
doi  ="10.1039/D4FD00093E",
url  ="http://dx.doi.org/10.1039/D4FD00093E"}

@article{segler2017ml-in-rs,
author = {Segler, Marwin H. S. and Waller, Mark P.},
title = {Neural-Symbolic Machine Learning for Retrosynthesis and Reaction Prediction},
journal = {Chemistry – A European Journal},
volume = {23},
number = {25},
pages = {5966-5971},
doi = {https://doi.org/10.1002/chem.201605499},
url = {https://chemistry-europe.onlinelibrary.wiley.com/doi/abs/10.1002/chem.201605499},
eprint = {https://chemistry-europe.onlinelibrary.wiley.com/doi/pdf/10.1002/chem.201605499},
year = {2017}
}

@ARTICLE{segler2018-solvability-mcts,
  title     = "Planning chemical syntheses with deep neural networks and
               symbolic {AI}",
  author    = "Segler, Marwin H S and Preuss, Mike and Waller, Mark P",
  journal   = "Nature",
  publisher = "Springer Science and Business Media LLC",
  volume    =  555,
  number    =  7698,
  pages     = "604--610",
  month     =  mar,
  year      =  2018,
  language  = "en",
  url ={https://www.nature.com/articles/nature25978},
  doi = {10.1038/nature25978}
}

@misc{deepseekai2025deepseekv32,
      title={DeepSeek-V3.2: Pushing the Frontier of Open Large Language Models}, 
      author={DeepSeek-AI},
      year={2025},
      eprint={2512.02556},
      archivePrefix={arXiv},
      primaryClass={cs.CL},
      url={https://arxiv.org/abs/2512.02556}, 
}

@misc{kimiteam2026kimik25visualagentic,
      title={Kimi K2.5: Visual Agentic Intelligence}, 
      author={Kimi-Team},
      year={2026},
      eprint={2602.02276},
      archivePrefix={arXiv},
      primaryClass={cs.CL},
      url={https://arxiv.org/abs/2602.02276}, 
}

@misc{glm5team2026glm5vibecodingagentic,
      title={GLM-5: from Vibe Coding to Agentic Engineering}, 
      author={GLM-5-Team},
      year={2026},
      eprint={2602.15763},
      archivePrefix={arXiv},
      primaryClass={cs.LG},
      url={https://arxiv.org/abs/2602.15763}, 
}

@misc{shao2024deepseekmathpushinglimitsmathematical,
      title={DeepSeekMath: Pushing the Limits of Mathematical Reasoning in Open Language Models}, 
      author={Zhihong Shao and Peiyi Wang and Qihao Zhu and Runxin Xu and Junxiao Song and Xiao Bi and Haowei Zhang and Mingchuan Zhang and Y. K. Li and Y. Wu and Daya Guo},
      year={2024},
      eprint={2402.03300},
      archivePrefix={arXiv},
      primaryClass={cs.CL},
      url={https://arxiv.org/abs/2402.03300}, 
}

@article{Lowe2017usptofull,
author = "Daniel Lowe",
title = "{Chemical reactions from US patents (1976-Sep2016)}",
year = "2017",
month = "6",
url = "https://figshare.com/articles/dataset/Chemical_reactions_from_US_patents_1976-Sep2016_/5104873",
doi = "10.6084/m9.figshare.5104873.v1"
}

@misc{pistachio_nextmove,
  author       = {{NextMove Software}},
  title        = {Pistachio},
  howpublished = {\url{https://www.nextmovesoftware.com/pistachio.html}},
  note         = {Commercial reaction database}
}

@misc{xai2025grok41,
  author = {xAI},
  title = {Grok 4.1 Model Card},
  year = {2025},
  month = {November},
  url = {https://data.x.ai/2025-11-17-grok-4-1-model-card.pdf},
}

@misc{xai2026grok43,
  author = {xAI},
  title  = {Grok 4.3},
  year   = {2026},
  howpublished = {xAI API Documentation},
  url    = {https://docs.x.ai/developers/models},
}

@misc{openai2025gpt51,
  author = {OpenAI},
  title = {GPT-5.1 Instant and GPT-5.1 Thinking System Card Addendum},
  year = {2025},
  month = {November},
  url = {https://cdn.openai.com/pdf/4173ec8d-1229-47db-96de-06d87147e07e/5_1_system_card.pdf},
}

@misc{openai2025gpt52,
  author = {OpenAI},
  title = {Update to GPT-5 System Card: GPT-5.2},
  year = {2025},
  month = {December},
  url = {https://cdn.openai.com/pdf/3a4153c8-c748-4b71-8e31-aecbde944f8d/oai_5_2_system-card.pdf},
}

@misc{openai2025gpt54,
  author = {OpenAI},
  title = {Introducing GPT‑5.4},
  year = {2026},
  month = {March},
  url = {https://openai.com/index/introducing-gpt-5-4/},
}

@misc{openai2025gpt55,
  author = {OpenAI},
  title = {Introducing GPT‑5.5},
  year = {2026},
  month = {April},
  url = {https://openai.com/index/introducing-gpt-5-5/},
}

@misc{anthropic2025claudesonnet45,
  author = {Anthropic},
  title = {System Card: Claude Sonnet 4.5},
  year = {2025},
  month = {September},
  url = {https://www.anthropic.com/claude-sonnet-4-5-system-card},
}

@misc{anthropic2025claudesonnet46,
  author = {Anthropic},
  title = {System Card: Claude Sonnet 4.6},
  year = {2026},
  month = {February},
  url = {https://www.anthropic.com/claude-sonnet-4-6-system-card},
}

@misc{anthropic2025claudeopus45,
  author = {Anthropic},
  title = {System Card: Claude Opus 4.5},
  year = {2025},
  month = {November},
  url = {https://www.anthropic.com/claude-opus-4-5-system-card},
}

@misc{anthropic2025claudeopus46,
  author = {Anthropic},
  title = {System Card: Claude Opus 4.6},
  year = {2026},
  month = {February},
  url = {https://www.anthropic.com/claude-opus-4-6-system-card},
}

@misc{anthropic2025claudeopus47,
  author = {Anthropic},
  title = {System Card: Claude Opus 4.7},
  year = {2026},
  month = {April},
  url = {https://www.anthropic.com/claude-opus-4-7-system-card},
}

@misc{anthropic2026claudeopus48,
  author = {Anthropic},
  title = {System Card: Claude Opus 4.8},
  year = {2026},
  month = {May},
  url = {https://www.anthropic.com/claude-opus-4-8-system-card},
}

@misc{QwenTeam2026qwen35,
  author = {Qwen-Team},
  title = {Model Card: Qwen3.5},
  year = {2026},
  month = {March},
  url = {https://qwen.ai/blog?id=qwen3.5},
}

@misc{google2026gemini3.1pro,
  author = {Gemini-Team},
  title = {Gemini 3.1 Pro },
  year = {2026},
  month = {February},
  url = {https://deepmind.google/models/model-cards/gemini-3-1-pro/},
}

@misc{amini2025lfm2,
      title={LFM2 Technical Report}, 
      author={Alexander Amini and Anna Banaszak and Harold Benoit and Arthur Böök and Tarek Dakhran and Song Duong and Alfred Eng and Fernando Fernandes and Marc Härkönen and Anne Harrington and Ramin Hasani and Saniya Karwa and Yuri Khrustalev and Maxime Labonne and Mathias Lechner and Valentine Lechner and Simon Lee and Zetian Li and Noel Loo and Jacob Marks and Edoardo Mosca and Samuel J. Paech and Paul Pak and Rom N. Parnichkun and Alex Quach and Ryan Rogers and Daniela Rus and Nayan Saxena and Bettina Schlager and Tim Seyde and Jimmy T. H. Smith and Aditya Tadimeti and Neehal Tumma},
      year={2025},
      eprint={2511.23404},
      archivePrefix={arXiv},
      primaryClass={cs.LG},
      url={https://arxiv.org/abs/2511.23404}, 
}

@misc{xie2023retrosknn,
      title={Retrosynthesis Prediction with Local Template Retrieval}, 
      author={Shufang Xie and Rui Yan and Junliang Guo and Yingce Xia and Lijun Wu and Tao Qin},
      year={2023},
      eprint={2306.04123},
      archivePrefix={arXiv},
      primaryClass={cs.AI},
      url={https://arxiv.org/abs/2306.04123}, 
}

@article{VLEDUTS1963117,
title = {Concerning one system of classification and codification of organic reactions},
journal = {Information Storage and Retrieval},
volume = {1},
number = {2},
pages = {117-146},
year = {1963},
issn = {0020-0271},
doi = {https://doi.org/10.1016/0020-0271(63)90013-5},
url = {https://www.sciencedirect.com/science/article/pii/0020027163900135},
author = {G.É. Vléduts}
}

@MISC{smarts,
  title        = "{SMARTS} — A Language for Describing Molecular Patterns",
  author    = "Daylight~Chemical~Information~Systems,~Inc.",
  year         =  2007,
  url = {https://www.daylight.com/dayhtml/doc/theory/theory.smarts.html}
}

@ARTICLE{Schwaller2026-synthegy,
  title     = "Chemical reasoning in {LLMs} unlocks strategy-aware synthesis
               planning and reaction mechanism elucidation",
  author    = "Bran, Andres M and Neukomm, Théo A and Armstrong, Daniel and
               Jončev, Zlatko and Schwaller, Philippe",
  journal   = "Matter",
  publisher = "Elsevier BV",
  volume    =  0,
  number    =  102812,
  pages     =  102812,
  month     =  apr,
  year      =  2026,
  language  = "en",
  doi={10.1016/j.matt.2026.102812},
  url={https://www.cell.com/matter/fulltext/S2590-2385(26)00175-X},
}

@ARTICLE{Chen2021localretro,
  title     = "Deep retrosynthetic reaction prediction using local reactivity
               and global attention",
  author    = "Chen, Shuan and Jung, Yousung",
  journal   = "JACS Au",
  publisher = "American Chemical Society (ACS)",
  volume    =  1,
  number    =  10,
  pages     = "1612--1620",
  month     =  oct,
  year      =  2021,
  language  = "en",
  doi = {10.1021/jacsau.1c00246},
  url = {https://pubs.acs.org/doi/10.1021/jacsau.1c00246}
}

@misc{tu2025askcos-buyables,
      title={ASKCOS: an open source software suite for synthesis planning}, 
      author={Zhengkai Tu and Sourabh J. Choure and Mun Hong Fong and Jihye Roh and Itai Levin and Kevin Yu and Joonyoung F. Joung and Nathan Morgan and Shih-Cheng Li and Xiaoqi Sun and Huiqian Lin and Mark Murnin and Jordan P. Liles and Thomas J. Struble and Michael E. Fortunato and Mengjie Liu and William H. Green and Klavs F. Jensen and Connor W. Coley},
      year={2025},
      eprint={2501.01835},
      archivePrefix={arXiv},
      primaryClass={cs.AI},
      url={https://arxiv.org/abs/2501.01835}, 
}

@ARTICLE{Zhong2022-root-aligned-smiles,
  title     = "Root-aligned {SMILES}: a tight representation for chemical
               reaction prediction",
  author    = "Zhong, Zipeng and Song, Jie and Feng, Zunlei and Liu, Tiantao and
               Jia, Lingxiang and Yao, Shaolun and Wu, Min and Hou, Tingjun and
               Song, Mingli",
  journal   = "Chem. Sci.",
  publisher = "Royal Society of Chemistry (RSC)",
  volume    =  13,
  number    =  31,
  pages     = "9023--9034",
  month     =  aug,
  year      =  2022,
  language  = "en",
  url = "https://pubs.rsc.org/en/content/articlelanding/2022/sc/d2sc02763a",
}

@article{sacha2021megan,
author = {Sacha, Mikołaj and Błaż, Mikołaj and Byrski, Piotr and Dąbrowski-Tumański, Paweł and Chromiński, Mikołaj and Loska, Rafał and Włodarczyk-Pruszyński, Paweł and Jastrzębski, Stanisław},
title = {Molecule Edit Graph Attention Network: Modeling Chemical Reactions as Sequences of Graph Edits},
journal = {Journal of Chemical Information and Modeling},
volume = {61},
number = {7},
pages = {3273-3284},
year = {2021},
doi = {10.1021/acs.jcim.1c00537},
note ={PMID: 34251814},
URL = {https://doi.org/10.1021/acs.jcim.1c00537},
eprint = {https://doi.org/10.1021/acs.jcim.1c00537}
}

@ARTICLE{Zhong2023-graph2edits,
  title     = "Retrosynthesis prediction using an end-to-end graph generative
               architecture for molecular graph editing",
  author    = "Zhong, Weihe and Yang, Ziduo and Chen, Calvin Yu-Chian",
  journal   = "Nat. Commun.",
  publisher = "Nature Publishing Group",
  volume    =  14,
  number    =  1,
  pages     =  3009,
  month     =  may,
  year      =  2023,
  language  = "en",
  url = "https://www.nature.com/articles/s41467-023-38851-5"
}

@inproceedings{coley_gln,
 author = {Dai, Hanjun and Li, Chengtao and Coley, Connor and Dai, Bo and Song, Le},
 booktitle = {Advances in Neural Information Processing Systems},
 editor = {H. Wallach and H. Larochelle and A. Beygelzimer and F. d\textquotesingle Alch\'{e}-Buc and E. Fox and R. Garnett},
 pages = {},
 publisher = {Curran Associates, Inc.},
 title = {Retrosynthesis Prediction with Conditional Graph Logic Network},
 url = {https://proceedings.neurips.cc/paper_files/paper/2019/file/0d2b2061826a5df3221116a5085a6052-Paper.pdf},
 volume = {32},
 year = {2019}
}

@ARTICLE{Seidl2022mhnreact,
  title     = "Improving few- and zero-shot reaction template prediction using
               Modern Hopfield Networks",
  author    = "Seidl, Philipp and Renz, Philipp and Dyubankova, Natalia and
               Neves, Paulo and Verhoeven, Jonas and Wegner, Jörg K and Segler,
               Marwin and Hochreiter, Sepp and Klambauer, Günter",
  journal   = "J. Chem. Inf. Model.",
  publisher = "American Chemical Society (ACS)",
  volume    =  62,
  number    =  9,
  pages     = "2111--2120",
  month     =  may,
  year      =  2022,
  language  = "en",
  url = "https://pubs.acs.org/doi/full/10.1021/acs.jcim.1c01065"
}

@inproceedings{zagribelnyy2026chemcensor_c3lm,
  title     = {When Single Answer Is Not Enough: Rethinking Single-Step Retrosynthesis Benchmarks for {LLMs}},
  author    = {Zagribelnyy, Bogdan and Ilin, Ivan and Kuznetsov, Maksim and Bondarev, Nikita and Schutski, Roman and MacDougall, Thomas and Shayakhmetov, Rim and Miftakhutdinov, Zulfat and Mizera, Mikolaj and Aladinskiy, Vladimir and Aliper, Alex and Zhavoronkov, Alex},
  booktitle = {Proceedings of the 43rd International Conference on Machine Learning},
  series    = {Proceedings of Machine Learning Research},
  volume    = {306},
  year      = {2026},
  publisher = {PMLR},
  address   = {Seoul, South Korea},
  url       = {https://arxiv.org/abs/2602.03554},
  note      = {Copy available here: \url{https://arxiv.org/abs/2602.03554}}
}

@article{chembl,
    author = {Zdrazil, Barbara and Felix, Eloy and Hunter, Fiona and Manners, Emma J and Blackshaw, James and Corbett, Sybilla and de Veij, Marleen and Ioannidis, Harris and Lopez, David Mendez and Mosquera, Juan F and Magarinos, Maria Paula and Bosc, Nicolas and Arcila, Ricardo and Kizilören, Tevfik and Gaulton, Anna and Bento, A Patrícia and Adasme, Melissa F and Monecke, Peter and Landrum, Gregory A and Leach, Andrew R},
    title = {The ChEMBL Database in 2023: a drug discovery platform spanning multiple bioactivity data types and time periods},
    journal = {Nucleic Acids Research},
    volume = {52},
    number = {D1},
    pages = {D1180-D1192},
    year = {2024},
    month = {01},
    issn = {0305-1048},
    doi = {10.1093/nar/gkad1004},
    url = {https://doi.org/10.1093/nar/gkad1004},
    eprint = {https://academic.oup.com/nar/article-pdf/52/D1/D1180/55040046/gkad1004.pdf},
}

@misc{zagribelnyy2026rersa,
  author       = {Zagribelnyy, Bogdan and Fedorchenko, Sergei and Bondarev, Nikita and Ilin, Ivan and Ivanenkov, Yan and Zavoronkovs, Aleksandrs},
  title        = {Advanced Retrosynthesis-Related Synthetic Accessibility Modeling},
  howpublished = {U.S. Patent Application Publication No. US 2026/0100252 A1},
  year         = {2026},
  month        = apr,
  note         = {Appl. No. 19/353,038; filed 2025-10-08; priority 2024-10-09},
  url          = {https://patents.google.com/patent/US20260100252A1/en}
}

@ARTICLE{Boiko2023-autonomous-llms,
  title    = "Autonomous chemical research with large language models",
  author   = "Boiko, Daniil A and MacKnight, Robert and Kline, Ben and Gomes,
              Gabe",
  journal  = "Nature",
  volume   =  624,
  number   =  7992,
  pages    = "570--578",
  month    =  dec,
  year     =  2023,
  language = "en",
  doi = {https://doi.org/10.1038/s41586-023-06792-0},
  url = {https://www.nature.com/articles/s41586-023-06792-0}
}

\appendix
\onecolumn

\clearpage
\section{C3LM Family Inventory}
\label{app:inventory}

\begin{table*}[h]
\centering
\begin{tabular}{@{}l|l|ll@{}}
\toprule
\textbf{Model Name} & \makecell[l]{\textbf{Short}\\\textbf{Name}} &  \makecell[l]{\textbf{Training}\\\textbf{Data}} & \textbf{Reward} \\
\midrule
\multicolumn{4}{c}{\textit{\ourmodel{}}, Supervised Fine-Tuning, \textbf{Top-1 task}} \\ 
\midrule
\ourmodel{}-LFM2-CREED-CCV+USPTO\textsuperscript{*}         & C3LM-1 & TD-1 & -- \\
\midrule
\multicolumn{4}{c}{\textit{\ourmodel{}}, Supervised Fine-Tuning, \textbf{Top-K task}} \\ 
\midrule
\ourmodel{}-LFM2-CREED-CCV+USPTO                            & C3LM-2 & TD-1 & -- \\
\ourmodel{}-LFM2-CREED-CCV-2+USPTO-XL                       & C3LM-3 & TD-2 & -- \\
\midrule
\multicolumn{4}{c}{\textit{\ourmodel{}}, Reinforcement Learning Fine-Tuning, \textbf{Top-K task}} \\
\midrule
\ourmodel{}-LFM2-RFT-CC                               & C3LM-4 & TD-2 & ChemCensor \\
\ourmodel{}-LFM2-RFT-CC-NR                            & C3LM-5 & TD-2 & ChemCensor, Novelty \\
\bottomrule
\end{tabular}
\caption{C3LM family inventory. \textbf{TD-1} = CREED-CCV + USPTO; \textbf{TD-2} = CREED-CCV-2 + USPTO-XL.
$^{*}$Trained in prior work~\citep{zagribelnyy2026chemcensor_c3lm}; its predictions are re-scored under ChemCensor (v1.1.1) for comparison, without retraining.}
\label{tab:inventory}
\end{table*}
\section{Glossary}
\label{app:gloassary}

\begin{description}

\item[\textbf{ChemCensor}]
ChemCensor Score (CC Score) is a quantitative metric evaluating the chemical plausibility of a reaction and/or a retrosynthetic route by measuring the proportion of steps that pass plausibility validation, weighted by their confidence levels.

\item[\textbf{CREED}]
CREED (Comprehensive Reactant Exhaustive Enumeration Dataset) is a large-scale, quality-controlled dataset of $\sim$22.7M unique reactions over 1{,}493{,}715 unique products introduced in \citep{zagribelnyy2026chemcensor_c3lm}, generated by a virtual synthesis engine from $\sim$3K expert-coded bidirectional reaction templates and designed to expose multiple plausible single-step disconnections per product rather than a single ground truth; its ChemCensor-verified subset, retaining only candidates with CC Score $> 0$, is denoted CREED-CCV.

\item[\textbf{C3LM}]
C3LM (Chemistry Constraint-Consistent Language Model) is the family of language models introduced in \citep{zagribelnyy2026chemcensor_c3lm}, fine-tuned on CREED and its variants from the LFM2 2.6B base checkpoint via supervised and reinforcement fine-tuning, and optimized to generate chemically plausible single-step retrosynthetic disconnections.

\item[\textbf{URSA-expert-2026}]
URSA-expert-2026 is an expert-annotated dataset introduced in \citep{zagribelnyy2026chemcensor_c3lm}, an out-of-distribution benchmark of 100 novel, machine-generated target molecules whose synthetic accessibility was confirmed by expert chemists, constructed to be disjoint from publicly available reaction datasets so as to evaluate SSRS generalization free from training-set memorization or data leakage.

\item[\textbf{USPTO-50K-test-mini}]
USPTO-50K-test-mini is a cost-efficient 10\% random subset (497 targets) of the curated USPTO-50K-test introduced in \citep{zagribelnyy2026chemcensor_c3lm}, recommended as the default USPTO-derived in-distribution benchmark for LLM-based SSRS evaluation under API-cost or wall-clock constraints.

\item[\textbf{Target molecule}]
The target molecule is the desired chemical compound that represents the ultimate goal of synthesis planning. It is the molecule for which the system generates or evaluates synthetic routes, and it serves as the starting point for retrosynthetic disconnection, working backward from the target to identify precursor molecules.

\item[\textbf{Reactants}]
Reactants are the chemical compounds that undergo transformation during a chemical reaction and whose atoms are directly incorporated into the product structure. Technically, reactants are distinguished from reagents by the presence of atom mapping -- atoms in reactants have corresponding mapped atoms in the product(s).

\item[\textbf{Reagents}]
Reagents are chemical compounds that participate in a chemical reaction but whose atoms are not directly incorporated into the product structure. Reagents typically facilitate or enable the transformation (e.g., catalysts, bases, acids, solvents with reactive roles) with no or only minor atom contribution to the final product. For benchmarking, it is reasonable to extend the reacting species with reagents, since they may influence the correctness of atom--atom mapping and thereby the reaction-center extraction process.

\item[\textbf{Starting materials}]
Starting materials are the initial chemical compounds from which a synthetic route begins. They are molecules that exist at the terminal nodes (leaves) of a retrosynthetic tree and are not produced by any reaction step within the route. Starting materials serve as the input chemicals for the synthesis and are expected to be commercially available or otherwise accessible within the reported synthetic methods.

\item[\textbf{Building blocks}]
Building blocks (CABBs) are commercially available chemical compounds that can be found in vendor datasets and purchased from them.

\item[\textbf{Single-step retrosynthesis model}]
An SSRS model predicts one or several retrosynthetic disconnections by mapping a target product molecule to a set of precursor reactants corresponding to a single reaction step. The model does not perform recursive planning or multi-step route construction, focusing instead on identifying chemically plausible reactants.

\item[\textbf{Multi-step retrosynthesis model}]
An MSRS model is a system that applies retrosynthetic transformations (typically recursively) to decompose a target molecule into commercially available or otherwise accessible starting materials through a sequence of reaction steps.

\item[\textbf{Reaction center (RC)}]
An RC is the set of atoms (dynamic atoms) in one or more reactant molecules and the product molecule that undergoes change during a chemical transformation, including atoms/bonds that are formed, broken, created, destroyed, or whose connectivity, bond order, formal charge, or hybridization state differs between reactants and products.

\item[\textbf{Chemical plausibility}]
Chemical plausibility reflects alignment of a reaction with core principles of organic synthesis (e.g., chemoselectivity, regioselectivity, stereoselectivity). Operationally, it can be reduced to chemoinformatic concepts such as reaction centers, functional groups, their occurrence, and compatibility rules, potentially augmented with conditions (solvents, temperature, catalysts, auxiliary reagents). In this system, plausibility is assessed by comparing the reaction center and functional-group context against a reference dataset of verified transformations. If the extracted reaction center is absent from the reference library, the reaction is considered implausible; similarly, functional groups never observed for that reaction center negate plausibility. When both reaction-center and functional-group context are supported by precedents, the reaction is considered chemically plausible.

\item[\textbf{Level of confidence}]
The nominal degree of chemical plausibility is estimated via discrete levels of confidence (LC), depending on which reaction-center representation is matched among verified transformations. Higher LC indicates that a more specific (larger-context) reaction-center definition is supported, correlating with higher nominal plausibility and representativeness in terms of synthetic precedents. The LC value is used as the reaction score in ChemCensor.

\item[\textbf{Functional groups (FGs)}]
FGs are structural motifs that determine chemical reactivity and properties. In the present system, functional groups are represented as SMARTS patterns \cite{smarts} that can be matched to molecular structures via substructure search. Functional-group context annotated for each reaction center helps determine which patterns are tolerable for a transformation, supporting chemical plausibility.

\item[\textbf{Functional group (FG) signature}]
An FG signature is the ensemble of FGs present in reactant/product molecules that are not affected by the transformation. For a given reaction center, the signature is constructed by aggregating synthetic precedents from the reference dataset.

\item[\textbf{Reference dataset}]
A reference dataset is a collection of verified reaction transformations extracted from sources including patents (e.g., USPTO), articles, preprints, and ELNs. It may include metadata such as conditions and yield. In this system, the reference dataset is used to validate analyzed reactions via reaction-center matching and functional-group signature comparison.

\item[\textbf{Synthetic precedent}]
A synthetic precedent is an elementary synthetic fact of a successful chemical reaction recorded in a reference dataset.

\end{description}
\section{Baselines}
\label{app:baselines}

\textbf{Proprietary Foundation Models:} Grok 4.1 \citep{xai2025grok41} and 4.3 \citep{xai2026grok43}; Gemini 3.1 Pro \citep{google2026gemini3.1pro};  GPT 5.1 \citep{openai2025gpt51}; 5.2 \citep{openai2025gpt52}, 5.4 \citep{openai2025gpt54} and 5.5 \citep{openai2025gpt55}; Claude Sonnet 4.5 \citep{anthropic2025claudesonnet45} and 4.6 \citep{anthropic2025claudesonnet46}; Claude Opus 4.5 \citep{anthropic2025claudeopus45}, 4.6 \citep{anthropic2025claudeopus46}, 4.7 \citep{anthropic2025claudeopus47} and 4.8 \citep{anthropic2026claudeopus48}.

\textbf{Open-weight Foundation Models:} DeepSeek-V3.2 \citep{deepseekai2025deepseekv32}; Qwen3.5-397B-A17B (formally Qwen3.5) \citep{QwenTeam2026qwen35}, Kimi K2.5 \citep{kimiteam2026kimik25visualagentic} and GLM-5 \citep{glm5team2026glm5vibecodingagentic}.

\textbf{Conventional SSRS models:} LocalRetro \citep{Chen2021localretro}, R-SMILES \citep{Zhong2022-root-aligned-smiles}, MEGAN \citep{sacha2021megan}, Chemformer \citep{irwin2022chemformer}, Graph2Edits \citep{Zhong2023-graph2edits}, GLN \citep{coley_gln}, MHNreact \citep{Seidl2022mhnreact} and RetroKNN \citep{xie2023retrosknn}.
\section{C3LM Supervised Fine-Tuning Details}
\label{app:c3lm_sft}

\textbf{Training procedure: } On each training step, we utilize a total $524,288$ tokens context windows across all GPUs and pack multiple training sequences into each GPU available context window. 

\textbf{Training Data Preprocessing:} Similarly to other chemical language models \citep{livne2024nacho, pei2023biot5}, we extend the base vocabulary with SMILES format \citep{weininger1988smiles} specific tokens; this is aimed at isolating chemical tokens from natural-language tokens and providing a consistent representation of SMILES entities. During the training, we tokenize SMILES into specialized tokens always in model outputs, and with $0.5$ probability for user input; we adopt this strategy to align LFM2 base checkpoint chemical knowledge and new tokens. To improve chemical generalization, we augment SMILES entities during training by applying non-canonical random traversal.

\textbf{Reasoning:} We use the same basic reasoning scheme as~\citep{zagribelnyy2026chemcensor_c3lm}: each chain-of-thought example is prepended with a deterministic block listing the canonical SMILES of the input entities and their BRICS fragments~\cite{degen2008brics}.

\section{C3LM Reinforcement Learning Fine-Tuning Details}
\label{app:c3lm_rft}

\textbf{Training procedure: } Online Reinforcement Learning Fine-Tuning (RFT) of the \ourmodel{} (CREED-CCV-2+USPTO-XL) model is performed using single-reward Group Relative Policy Optimization (GRPO) \cite{shao2024deepseekmathpushinglimitsmathematical}. RFT uses the same \traindataset{} training split as SFT, with sampling temperature of $1$ and KL-regularization weight of $0.1$. The policy is trained over $1000$ training steps, with a learning rate of $10^{-6}$, a GRPO group size of $8$, and $64$ groups per step. The GRPO reward function is defined as a weighted sum combination of multiple reward components. The components and their weights are as follows:

\textbf{Thinking format } (weight $0.1$) verifies if the generated completion is correctly formatted, by returning $1$ for keeping the thinking in thinking tags, and $-1$ otherwise.

\textbf{Molecular syntax } (weight $0.5$) verifies if the answer is a valid SMILES string.

\textbf{ChemCensor score } (weight $1.0$) verifies, for a generated reaction, the chemical plausibility from ChemCensor. The plausibility is rescaled from $(0, 5)$ to $(0, 1)$. A reward of $-1$ is assigned when the generated reactant SMILES is invalid.

\textbf{Top-K uniqueness } (weight $0.2$) computes the proportion of uniquely generated solutions, by comparing the canonical version of the generated SMILES string of each solution. The range of this reward function is thus $(1/k, 1)$.

\textbf{Top-K matching } (weight $0.1$) verifies if the number of generated answers matches $k$, the number of requests answers.

\textbf{Novelty score } (weight $1.0$) provides a binary score for each generated reactant. This score is $1$ if the generated reactant does not exist in the exhaustive list of CREED and if its ChemCensor score is positive, which corresponds to the lowest degree of confidence for chemical plausibility. In doing so, we encourage the policy to generate plausible reactants outside of the training data. The individual scores' average is used as novelty reward.
\clearpage
\section{Full Plausibility-Based Top-K Evaluation}
\label{app:full_table}
{\centering
\captionsetup{hypcap=false}
\setlength{\tabcolsep}{5pt}
\renewcommand{\arraystretch}{0.95}
{\fontsize{11}{12}\selectfont
\begin{tabular}{@{}l|cccc|cccc@{}}
\toprule
\multirow{3}{*}{\textbf{Model}} &
\multicolumn{4}{c|}{\textbf{URSA-expert-2026}} &
\multicolumn{4}{c}{\textbf{USPTO-50K-test-mini}} \\
\cline{2-5}\cline{6-9}
& \multirow{2}{*}{\textbf{Max}} & \multicolumn{3}{c|}{\textbf{Av. PT-Top-K CC}}
& \multirow{2}{*}{\textbf{Max}} & \multicolumn{3}{c}{\textbf{Av. PT-Top-K CC}} \\
&  & \textbf{@3} & \textbf{@5} & \textbf{@10}
&  & \textbf{@3} & \textbf{@5} & \textbf{@10} \\
\midrule
\multicolumn{9}{c}{\textit{Proprietary Foundation Models}} \\
\midrule
Grok-4.1                   & 1.86 & 1.56 & 1.31 & 0.86 & 3.99 & 2.61 & 2.02 & 1.24\\
Grok-4.3                   & 1.74 & 1.41 & 1.13 & 0.66 & 3.99 & 2.58 & 1.95 & 1.15\\
Gemini 3.1 Pro             & 1.91 & 1.69 & 1.46 & 1.08 & 4.32 & 2.92 & 2.34 & 1.59 \\
GPT 5.1                    & 0.59 & 0.32 & 0.21 & 0.11 & 1.22 & 0.65 & 0.43 & 0.22 \\
GPT 5.2                    & 0.43 & 0.28 & 0.20 & 0.12 & 1.48 & 0.88 & 0.62 & 0.34 \\
GPT 5.4                    & 1.19 & 0.78 & 0.55 & 0.29 & 2.34 & 1.43 & 1.02 & 0.55 \\
GPT 5.5                    & 1.94 & 1.68 & 1.46 & 1.05 & 4.50 & 3.07 & 2.45 & 1.63 \\
Claude Sonnet 4.5          & 1.56 & 1.27 & 0.98 & 0.57 & 3.16 & 2.02 & 1.51 & 0.87 \\
Claude Sonnet 4.6          & 1.03 & 0.79 & 0.60 & 0.34 & 2.93 & 1.83 & 1.35 & 0.76 \\
Claude Opus 4.5            & 1.68 & 1.31 & 1.03 & 0.62 & 3.30 & 2.17 & 1.64 & 0.97 \\
Claude Opus 4.6            & 1.68 & 1.36 & 1.10 & 0.67 & 3.77 & 2.51 & 1.92 & 1.15 \\
Claude Opus 4.7            & 1.91 & 1.65 & 1.39 & 0.95 & 4.35 & 2.96 & 2.31 & 1.45 \\
Claude Opus 4.8            & 1.89 & 1.62 & 1.34 & 0.85 & 4.35 & 2.95 & 2.30 & 1.44 \\

\midrule
\multicolumn{9}{c}{\textit{Open-weight Foundation Models}} \\
\midrule
DeepSeek 3.2               & 0.50 & 0.34 & 0.24 & 0.13 & 0.95 & 0.56 & 0.38 & 0.21 \\
Qwen 3.5                   & 1.61 & 1.32 & 1.07 & 0.64 & 3.44 & 2.39 & 1.84 & 1.09\\
Kimi K2.5                  & 1.68 & 1.38 & 1.13 & 0.69 & 3.65 & 2.43 & 1.86 & 1.10\\
GLM-5                      & 1.22 & 0.97 & 0.75 & 0.42 & 2.16 & 1.39 & 1.03 & 0.58\\
\midrule
\multicolumn{9}{c}{\textit{Conventional SSRS Models}} \\
\midrule
LocalRetro          & 2.11 & \cellcolor{bronze!40}1.85 & 1.59 & 1.22 & 4.84 & 3.31 & \cellcolor{silver!40}2.67 & \cellcolor{bronze!40}1.81 \\
GLN                 & 1.96 & 1.72 & 1.49 & 1.03 & 4.80 & 3.18 & \cellcolor{bronze!40}2.52 & 1.58 \\
MEGAN               & 2.01 & 1.70 & 1.42 & 0.92 & 4.78 & 3.15 & 2.45 & 1.54 \\
Chemformer          & 1.77 & 1.03 & 0.68 & 0.35 & 4.67 & 1.69 & 1.02 & 0.51 \\
Graph2Edits         & 2.08 & 1.78 & 1.48 & 1.01 & 4.79 & 3.00 & 2.28 & 1.39 \\
MHNreact            & 2.05 & 1.84 & \cellcolor{bronze!40}1.62 & \cellcolor{bronze!40}1.28 & 4.86 & \cellcolor{bronze!40}3.30 & \cellcolor{gold!40}2.68 & \cellcolor{gold!40}1.90 \\
RetroKNN            & 2.10 & 1.84 & 1.60 & 1.22 & 4.85 & \cellcolor{silver!40}3.33 & \cellcolor{gold!40}2.68 & 1.82 \\
R-SMILES            & 2.08 & 1.83 & 1.56 & 1.11 & 4.85 & \cellcolor{gold!40}3.36 & \cellcolor{silver!40}2.67 & 1.75 \\
\midrule
\multicolumn{9}{c}{\textit{\ourmodel{}}, Supervised Fine-Tuning, \textbf{Top-1 Mode}} \\
\midrule
\ourmodel{}-LFM2-CREED-CCV+USPTO\textsuperscript{*}              & 1.62 & 1.06 & 0.72 & 0.38 & 4.12 & 2.10 & 1.36 & 0.70 \\
\midrule
\multicolumn{9}{c}{\textit{\ourmodel{}}, Supervised and Reinforcement Learning Fine-Tuning, \textbf{Top-K Mode}} \\
\midrule
\ourmodel{}-LFM2-CREED-CCV+USPTO         & 1.92 & 1.68 & 1.42 & 0.98 & 4.16 & 2.74 & 2.17 & 1.42 \\
\ourmodel{}-LFM2-CREED-CCV-2+USPTO-XL    & 2.04 & 1.81 & 1.59 & 1.27 & 4.16 & 2.88 & 2.37 & 1.70 \\ 
\ourmodel{}-LFM2-RFT-CC          & 2.08 & \cellcolor{silver!40}1.88 & \cellcolor{silver!40}1.65 & \cellcolor{silver!40}1.29 & 4.16 & 2.92 & 2.41 & 1.73 \\
\ourmodel{}-LFM2-RFT-CC-NR       & 2.16 & \cellcolor{gold!40}1.94 & \cellcolor{gold!40}1.73 & \cellcolor{gold!40}1.37 & 4.28 & 3.01 & 2.51 & \cellcolor{silver!40}1.85 \\
\midrule
\multicolumn{9}{c}{\textit{Chemical Plausibility and Diversity Frontier of Generated Reactions}} \\
\midrule
All GP LLMs (17) together                        & 2.19 & 2.07 & 1.93 & 1.65 & 4.85   & 3.98   & 3.53   & 2.87   \\
All conventional SSRS models (8) together             & 2.18 & 1.99 & 1.78 & 1.45 & 4.90   & 3.58   & 3.00   & 2.26   \\
All C3LMs (5) together              & 2.23   & 2.04   & 1.88   & 1.57   & 4.74   & 3.31   & 2.78   & 2.14   \\
All benchmarked models (30) together              & 2.25   & 2.15   & 2.04   & 1.81   & 4.91   & 4.12   & 3.69   & 3.04   \\
\bottomrule
\end{tabular}

\par}
\captionof{table}{Full plausibility-based evaluation in the single-step retrosynthesis Top-$K$ mode. \textbf{Max}: per-target maximum ChemCensor score averaged over TMs. \textbf{Av.\ PT-Top-K CC}: per-TM average ChemCensor score over top-K unique predictions; ChemCensor v1.1.1., USPTO-full as the source of synthetic precedents.}
\label{tab:main_table_topk_full}
\par}
\clearpage
\section{Plausibility-Based Top-1 (Single-Answer) Results}

The Top-1 (single-answer) predictions evaluated in \autoref{tab:top1} are taken from the original benchmark~\citep{zagribelnyy2026chemcensor_c3lm}; we do not regenerate them and only re-score them under ChemCensor v1.1.1, the version used throughout this paper, so that they are directly comparable with our Top-$K$ results.

\label{app:single_anwer}
\begin{table*}[h]
\centering
\begin{tabular}{@{}l|cccc|cccc@{}}
\toprule
\multirow{2}{*}{\textbf{Model}} &
\multicolumn{4}{c|}{\textbf{URSA-expert-2026}} &
\multicolumn{4}{c}{\textbf{USPTO-50K-test-mini}} \\
\cline{2-5}\cline{6-9}
& \multirow{2}{*}{\textbf{Max}} & \multicolumn{3}{c|}{\textbf{Av. PT-Top-K CC}}
& \multirow{2}{*}{\textbf{Max}} & \multicolumn{3}{c}{\textbf{Av. PT-Top-K CC}} \\
&  & \textbf{@3} & \textbf{@5} & \textbf{@10}
&  & \textbf{@3} & \textbf{@5} & \textbf{@10} \\
\midrule
\multicolumn{9}{c}{\textit{Proprietary Foundation Models}} \\
\midrule
Grok-4.1                   & 1.73 & 1.28 & 0.92 & 0.47 & 4.01   & 2.29   & 1.53   & 0.79   \\
Grok-4.3                   & 1.47 & 0.86 & 0.56 & 0.28 & 3.06   & 1.34   & 0.83   & 0.41   \\
Gemini 3.1 Pro             & 1.63 & 0.92 & 0.57 & 0.28 & 4.00   & 1.70   & 1.03   & 0.51   \\
GPT 5.1                    & 0.73 & 0.34 & 0.21 & 0.11 & 1.37   & 0.63   & 0.38   & 0.19   \\
GPT 5.2                    & 0.86 & 0.45 & 0.28 & 0.14 & 2.02   & 0.95   & 0.58   & 0.29   \\
GPT 5.4                    & 0.11 & 0.05 & 0.03 & 0.02 & 2.12   & 0.97   & 0.60   & 0.30   \\
GPT 5.5                    & 1.52 & 0.96 & 0.63 & 0.32 & 3.90   & 1.89   & 1.18   & 0.59   \\
Claude Sonnet 4.5          & 1.44 & 0.86 & 0.56 & 0.28 & 3.34   & 1.67   & 1.05   & 0.52   \\
Claude Sonnet 4.6          & 1.26 & 0.71 & 0.44 & 0.22 & 3.32   & 1.64   & 1.01   & 0.51   \\
Claude Opus 4.5            & 1.31 & 0.68 & 0.42 & 0.21 & 3.33   & 1.54   & 0.94   & 0.47   \\
Claude Opus 4.6            & 1.36 & 0.84 & 0.53 & 0.26 & 3.63   & 1.81   & 1.12   & 0.56   \\
Claude Opus 4.7            & 1.67 & 1.04 & 0.66 & 0.33 & 3.72   & 1.79   & 1.12   & 0.56   \\
Claude Opus 4.8            & 1.66 & 1.06 & 0.67 & 0.34 & 3.64   & 1.69   & 1.04   & 0.52   \\
\midrule
\multicolumn{9}{c}{\textit{Open-weight Foundation Models}} \\
\midrule
DeepSeek 3.2               & 0.39 & 0.16 & 0.09 & 0.05 & 1.14   & 0.44   & 0.26   & 0.13   \\
Qwen 3.5                   & 1.54 & 1.04 & 0.73 & 0.38 & 3.69   & 2.02   & 1.32   & 0.67   \\
Kimi K2.5                  & 1.47 & 0.98 & 0.64 & 0.33 & 3.73   & 1.85   & 1.16   & 0.58   \\
GLM-5                      & 1.03 & 0.49 & 0.29 & 0.15 & 3.67   & 1.92   & 1.23   & 0.62   \\
LFM2 2.6B                  & 0.00 & 0.00 & 0.00 & 0.00 & 0.00   & 0.00   & 0.00   & 0.00   \\
\bottomrule
\end{tabular}

\caption{Plausibility-based evaluation in the single-step retrosynthesis Top-$1$ mode. \textbf{Max}: per-target maximum ChemCensor score averaged over TMs. \textbf{Av.\ PT-Top-K CC}: per-TM average ChemCensor score over top-K unique predictions; ChemCensor v1.1.1.}
\label{tab:top1}
\end{table*}

\clearpage
\section{Reaction Intersection with the Conventional Model MHNreact}
\label{app:model_intersection}

To assess whether the models generate plausible reactions beyond those found by a strong conventional method, we compare the set of reactions predicted by each model against those predicted by MHNreact, the strongest conventional SSRS baseline in our evaluation. \autoref{tab:intersection} reports, for every model, the average per-target number of reactions in the intersection with MHNreact, those unique to the model, and those unique to MHNreact, together with the difference $\Delta$ between the number of reactions unique to the model and those unique to MHNreact.

\begin{table*}[h]
\centering
\begin{tabular}{lcccc}
  \hline
  \textbf{Model} & \textbf{Intersection} & \makecell[l]{\textbf{Unique} \\ \textbf{for Model}} & \makecell[l]{\textbf{Unique} \\ \textbf{for MHNreact}}  & \textbf{$\Delta$} \\
  \hline
  C3LM-LFM2-CREED-CCV-2+USPTO-XL & 5.0 & 6.8 & 6.4 & +0.4 \\
  C3LM-LFM2-RFT-CC-NR & 5.0 & 6.7 & 6.4 & +0.3 \\
  C3LM-LFM2-RFT-CC & 4.8 & 6.0 & 6.6 & -0.6 \\
  C3LM-LFM2-CREED-CCV+USPTO & 3.6 & 4.0 & 7.9 & -3.9 \\
  Gemini 3.1 Pro & 5.2 & 3.5 & 6.2 & -2.7 \\
  GPT 5.5 & 5.2 & 3.2 & 6.2 & -3.0 \\
  Grok-4.1 & 3.9 & 2.7 & 7.5 & -4.8 \\
  Claude Opus 4.7 & 4.5 & 2.6 & 6.9 & -4.3 \\
  Claude Opus 4.8 & 4.0 & 2.2 & 7.4 & -5.2 \\
  Qwen 3.5 & 2.7 & 2.0 & 8.7 & -6.7 \\
  Claude Opus 4.6 & 3.0 & 1.9 & 8.5 & -6.6 \\
  Grok-4.3 & 3.1 & 1.9 & 8.4 & -6.5 \\
  Kimi K2.5 & 3.1 & 1.9 & 8.4 & -6.5 \\
  Claude Sonnet 4.5 & 2.3 & 1.9 & 9.1 & -7.2 \\
  Claude Opus 4.5 & 2.7 & 1.8 & 8.7 & -6.9 \\
  GLM-5 & 2.3 & 1.4 & 9.2 & -7.8 \\
  C3LM-LFM2-CREED-CCV+USPTO* & 1.4 & 1.3 & 10.0 & -8.7 \\
  GPT 5.4 & 1.0 & 1.1 & 10.5 & -9.4 \\
  Claude Sonnet 4.6 & 1.6 & 0.8 & 9.8 & -9.0 \\
  GPT 5.2 & 0.3 & 0.6 & 11.1 & -10.5 \\
  GPT 5.1 & 0.3 & 0.5 & 11.1 & -10.6 \\
  DeepSeek 3.2 & 0.4 & 0.4 & 10.9 & -10.5 \\
  \hline
\end{tabular}

\caption{Intersection of predicted reactions for URSA-expert-2026 benchmark set with the reference conventional model MHNreact. $\Delta$ is the per-target difference between the number of reactions unique to the model and those unique to MHNreact.}
\label{tab:intersection}
\end{table*}

\begin{figure}[htbp]
  \centering
  \includegraphics[width=0.6\columnwidth]{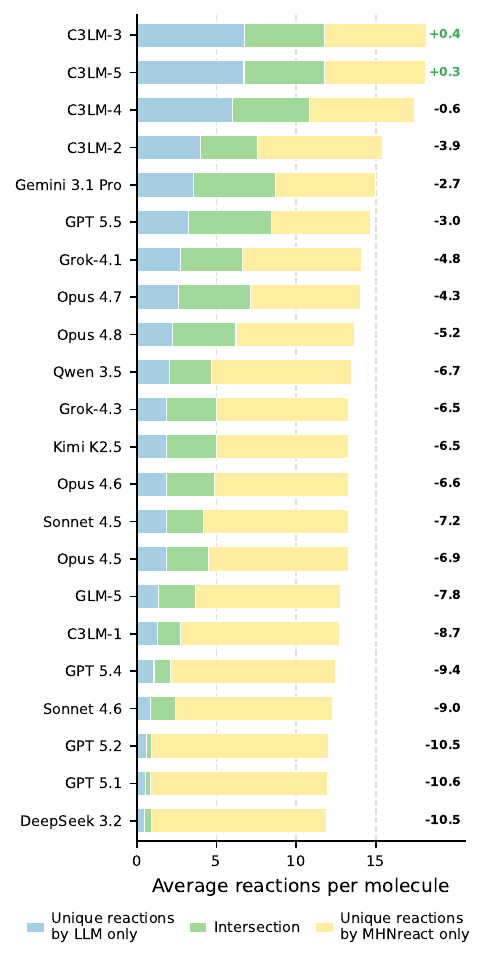}
  \caption{Intersection of reactions predicted by LLMs and the conventional SSRS model MHNreact. Values on the right show the difference between the number of reactions unique to the model and those unique to MHNreact (model $-$ MHNreact). \textbf{C3LM-1}: \texttt{C3LM-LFM2-CREED-CCV+USPTO\textsuperscript{*}};  \textbf{C3LM-2}: \texttt{C3LM-LFM2-CREED-CCV+USPTO}; \textbf{C3LM-3}: \texttt{C3LM-LFM2-CREED-CCV-2+USPTO-XL}; \textbf{C3LM-4}: \texttt{C3LM-LFM2-RFT-CC}; \textbf{C3LM-5}: \texttt{C3LM-LFM2-RFT-CC-NR}.}
  \label{fig:intersection_full}
\end{figure}

\autoref{fig:intersection_full} visualizes this partition across all models. Most models, including all foundation LLMs, produce fewer unique reactions than MHNreact ($\Delta < 0$); only the strongest C3LM variants --- \texttt{C3LM-LFM2-CREED-CCV-2+USPTO-XL} and \texttt{C3LM-LFM2-RFT-CC-NR} --- generate more unique plausible reactions than MHNreact ($\Delta > 0$), indicating that they extend the plausible reaction space beyond this conventional baseline.

\begin{figure}[t]
  \centering
  \includegraphics[width=\linewidth, trim=0 80 230 0, clip]{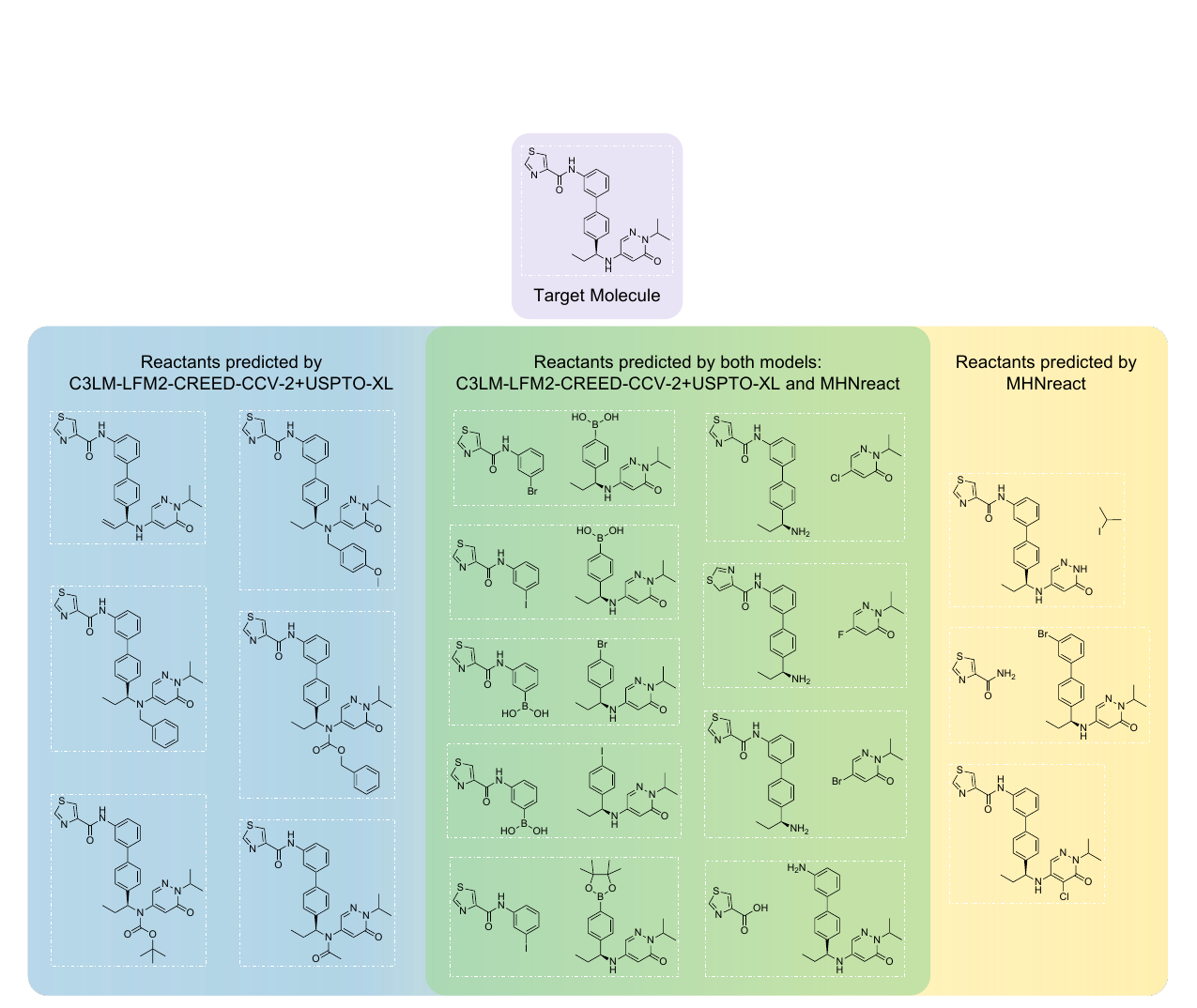}
  \caption{Intersection of reactions predicted for X404-1768-5005 by \texttt{C3LM-LFM2-CREED-CCV-2+USPTO-XL} and MHNreact.}
  \label{fig:intersection_example}
\end{figure}

Finally, \autoref{fig:intersection_example} gives a representative example for a single target (X404-1768-5005), contrasting the reactants predicted only by \texttt{C3LM-LFM2-CREED-CCV-2+USPTO-XL}, only by MHNreact, and by both models.

\clearpage
\section{ChemCensor Metrics Calculated with USPTO $\cup$ Pistachio Reference dataset}
\label{app:u2p2}

The plausibility scores in the main results (\autoref{tab:main_table_topk}) are computed by ChemCensor (v1.1.1) against its default reference database (USPTO-full, as reported in \citep{zagribelnyy2026chemcensor_c3lm}). To verify that our conclusions do not depend on this choice, we re-score the same model predictions with ChemCensor v1.1.1 using the ChemCensor-U2P2 reference database (USPTO-full $\cup$ Pistachio~Q3~2023) instead. As shown in \autoref{tab:main_table_topk_u2p2}, the absolute scores change, but the relative ordering of models is largely preserved, confirming that the comparison is robust to the choice of reference database.

\begin{table*}[th!]
\centering
\fontsize{10.5}{12}\selectfont{
\begin{tabular}{@{}l|cccc|cccc@{}}
\toprule
\multirow{3}{*}{\textbf{Model}} &
\multicolumn{4}{c|}{\textbf{URSA-expert-2026}} &
\multicolumn{4}{c}{\textbf{USPTO-50K-test-mini}} \\
\cline{2-5}\cline{6-9}
& \multirow{2}{*}{\textbf{Max}} & \multicolumn{3}{c|}{\textbf{Av. PT-Top-K CC}}
& \multirow{2}{*}{\textbf{Max}} & \multicolumn{3}{c}{\textbf{Av. PT-Top-K CC}} \\
&  & \textbf{@3} & \textbf{@5} & \textbf{@10}
&  & \textbf{@3} & \textbf{@5} & \textbf{@10} \\
\midrule
\multicolumn{9}{c}{\textit{Proprietary Foundation Models}} \\
\midrule
Grok-4.1                   & 2.17 & 1.81 & 1.54 & 1.06 & 4.13 & 2.84 & 2.27 & 1.46\\
Grok-4.3                   & 2.02 & 1.62 & 1.32 & 0.82 & 4.14 & 2.81 & 2.19 & 1.35\\
Gemini 3.1 Pro             & 2.20 & 1.93 & 1.68 & 1.29 & 4.42 & 3.12 & 2.56 & 1.83 \\
GPT 5.1                    & 0.69 & 0.40 & 0.27 & 0.14 & 1.41 & 0.79 & 0.54 & 0.28 \\
GPT 5.2                    & 0.51 & 0.33 & 0.24 & 0.15 & 1.61 & 0.99 & 0.71 & 0.40 \\
GPT 5.4                    & 1.37 & 0.97 & 0.71 & 0.39 & 2.50 & 1.62 & 1.19 & 0.66 \\
GPT 5.5                    & 2.28 & 1.94 & 1.68 & 1.27 & 4.60 & 3.26 & 2.68 & 1.86 \\
Claude Sonnet 4.5          & 1.86 & 1.51 & 1.21 & 0.75 & 3.36 & 2.31 & 1.80 & 1.11 \\
Claude Sonnet 4.6          & 1.17 & 0.90 & 0.70 & 0.42 & 3.13 & 2.07 & 1.57 & 0.92 \\
Claude Opus 4.5            & 1.92 & 1.55 & 1.24 & 0.79 & 3.54 & 2.46 & 1.94 & 1.22 \\
Claude Opus 4.6            & 1.95 & 1.59 & 1.29 & 0.83 & 3.96 & 2.77 & 2.19 & 1.37 \\
Claude Opus 4.7            & 2.24 & 1.89 & 1.61 & 1.16 & 4.44 & 3.16 & 2.55 & 1.68 \\
Claude Opus 4.8            & 2.17 & 1.85 & 1.54 & 1.02 & 4.45 & 3.17 & 2.55 & 1.68 \\

\midrule
\multicolumn{9}{c}{\textit{Open-weight Foundation Models}} \\
\midrule
DeepSeek 3.2               & 0.55 & 0.40 & 0.29 & 0.17 & 1.18 & 0.73 & 0.52 & 0.30 \\
Qwen 3.5                   & 1.91 & 1.54 & 1.26 & 0.81 & 3.61 & 2.61 & 2.07 & 1.29\\
Kimi K2.5                  & 1.98 & 1.61 & 1.35 & 0.88 & 3.84 & 2.67 & 2.11 & 1.33\\
GLM-5                      & 1.76 & 1.40 & 1.11 & 0.66 & 3.12 & 2.11 & 1.61 & 0.96\\
\midrule
\multicolumn{9}{c}{\textit{Conventional SSRS Models}} \\
\midrule
LocalRetro          & 2.38 & 2.11 & 1.84 & 1.41 & 4.88 & 3.49 & 2.87 & 2.02 \\
GLN                 & 2.26 & 1.94 & 1.68 & 1.21 & 4.86 & 3.35 & 2.71 & 1.76 \\
MEGAN               & 2.34 & 1.95 & 1.61 & 1.09 & 4.85 & 3.34 & 2.67 & 1.72 \\
Chemformer          & 2.01 & 1.16 & 0.78 & 0.40 & 4.73 & 1.73 & 1.05 & 0.52 \\
Graph2Edits         & 2.38 & 2.05 & 1.73 & 1.22 & 4.86 & 3.18 & 2.46 & 1.55 \\
MHNreact            & 2.33 & 2.08 & 1.82 & 1.44 & 4.90 & 3.48 & 2.89 & 2.12 \\
RetroKNN            & 2.35 & 2.11 & 1.82 & 1.41 & 4.90 & 3.53 & 2.91 & 2.04 \\
R-SMILES            & 2.35 & 2.09 & 1.79 & 1.30 & 4.90 & 3.54 & 2.89 & 1.95 \\
\midrule
\multicolumn{9}{c}{\textit{\ourmodel{}}, Supervised Fine-Tuning, \textbf{Top-1 Mode}} \\
\midrule
\ourmodel{}-LFM2-CREED-CCV+USPTO\textsuperscript{*}              & 1.88 & 1.24 & 0.86 & 0.45 & 4.23 & 2.23 & 1.46 & 0.76 \\
\midrule
\multicolumn{9}{c}{\textit{\ourmodel{}}, Supervised and Reinforcement Learning Fine-Tuning, \textbf{Top-K Mode}} \\
\midrule
\ourmodel{}-LFM2-CREED-CCV+USPTO         & 2.18 & 1.89 & 1.62 & 1.13 & 4.27 & 2.94 & 2.37 & 1.60 \\
\ourmodel{}-LFM2-CREED-CCV-2+USPTO-XL    & 2.32 & 2.05 & 1.80 & 1.43 & 4.29 & 3.08 & 2.58 & 1.91 \\
\ourmodel{}-LFM2-RFT-CC     & 2.42   & 2.14   & 1.89   & 1.48   & 4.30   & 3.12   & 2.62   & 1.93   \\
\ourmodel{}-LFM2-RFT-CC-NR  & 2.42   & 2.21   & 1.95   & 1.56   & 4.40   & 3.21   & 2.71   & 2.06   \\
\bottomrule
\end{tabular}

}
\caption{Plausibility-based evaluation in the single-step retrosynthesis Top-$K$ mode. \textbf{Max}: per-target maximum ChemCensor score averaged over TMs. \textbf{Av.\ PT-Top-K CC}: per-TM average ChemCensor score over top-K unique predictions; ChemCensor v1.1.1, \textbf{ChemCensor-U2P2 reference database (USPTO-full $\cup$ Pistachio~Q3~2023)}.}
\label{tab:main_table_topk_u2p2}
\end{table*}

\clearpage
\clearpage
\section{Results under ChemCensor v0.5.2}
\label{app:rescore-052}

The main results (\autoref{tab:main_table_topk}) are computed with ChemCensor v1.1.1. To verify that our conclusions do not depend on the specific version of the plausibility metric, we re-score the same model predictions with ChemCensor v0.5.2, the version used in the original benchmark~\citep{zagribelnyy2026chemcensor_c3lm}. The evaluation setup is otherwise identical to \autoref{tab:main_table_topk}; only the scoring function differs. As shown in \autoref{tab:main_table_topk_052}, the absolute values shift slightly, but the relative ordering of models is largely preserved, indicating that the comparison is robust to the choice of ChemCensor version.

\begin{table*}[th!]
\centering
\fontsize{10.5}{12}\selectfont{
\begin{tabular}{@{}l|cccc|cccc@{}}
\toprule
\multirow{3}{*}{\textbf{Model}} &
\multicolumn{4}{c|}{\textbf{URSA-expert-2026}} &
\multicolumn{4}{c}{\textbf{USPTO-50K-test-mini}} \\
\cline{2-5}\cline{6-9}
& \multirow{2}{*}{\textbf{Max}} & \multicolumn{3}{c|}{\textbf{Av. PT-Top-K CC}}
& \multirow{2}{*}{\textbf{Max}} & \multicolumn{3}{c}{\textbf{Av. PT-Top-K CC}} \\
&  & \textbf{@3} & \textbf{@5} & \textbf{@10}
&  & \textbf{@3} & \textbf{@5} & \textbf{@10} \\
\midrule
\multicolumn{9}{c}{\textit{Proprietary Foundation Models}} \\
\midrule
Grok-4.1                & 1.90 & 1.59 & 1.33 & 0.88 & 4.04 & 2.71 & 2.13 & 1.33 \\
Grok-4.3                & 1.80 & 1.45 & 1.16 & 0.69 & 4.02 & 2.65 & 2.03 & 1.22 \\
Gemini 3.1 Pro          & 1.96 & 1.72 & 1.49 & 1.11 & 4.35 & 2.97 & 2.42 & 1.66 \\
GPT 5.1                 & 0.61 & 0.35 & 0.23 & 0.12 & 1.31 & 0.72 & 0.48 & 0.25 \\
GPT 5.2                 & 0.43 & 0.28 & 0.20 & 0.12 & 1.53 & 0.92 & 0.65 & 0.36 \\
GPT 5.4                 & 1.23 & 0.81 & 0.57 & 0.30 & 2.40 & 1.50 & 1.08 & 0.59 \\
GPT 5.5                 & 1.98 & 1.71 & 1.48 & 1.08 & 4.51 & 3.15 & 2.55 & 1.72 \\
Claude Sonnet 4.5       & 1.59 & 1.29 & 1.02 & 0.60 & 3.21 & 2.10 & 1.58 & 0.93 \\
Claude Sonnet 4.6      & 1.04 & 0.79 & 0.61 & 0.35 & 2.98 & 1.92 & 1.42 & 0.81 \\
Claude Opus 4.5         & 1.70 & 1.34 & 1.07 & 0.65 & 3.37 & 2.28 & 1.75 & 1.06 \\
Claude Opus 4.6         & 1.73 & 1.39 & 1.13 & 0.70 & 3.82 & 2.61 & 2.02 & 1.23 \\
Claude Opus 4.7         & 1.95 & 1.68 & 1.42 & 0.97 & 4.37 & 3.04 & 2.40 & 1.53 \\
Claude Opus 4.8         & 1.93 & 1.66 & 1.38 & 0.88 & 4.37 & 3.03 & 2.40 & 1.51 \\
\midrule
\multicolumn{9}{c}{\textit{Open-weight Foundation Models}} \\
\midrule
DeepSeek 3.2            & 0.51 & 0.35 & 0.24 & 0.13 & 0.99 & 0.60 & 0.42 & 0.23 \\
Qwen 3.5               & 1.63 & 1.33 & 1.07 & 0.64 & 3.49 & 2.46 & 1.92 & 1.16 \\
Kimi K2.5              & 1.73 & 1.42 & 1.16 & 0.71 & 3.72 & 2.51 & 1.94 & 1.17 \\
GLM-5                  & 1.50 & 1.20 & 0.93 & 0.52 & 2.93 & 1.92 & 1.44 & 0.83 \\
\midrule
\multicolumn{9}{c}{\textit{Conventional SSRS Models}} \\
\midrule
LocalRetro  & 2.14 & 1.87 & 1.61 & 1.23 & 4.82 & 3.35 & 2.73 & 1.88 \\
GLN         & 1.97 & 1.73 & 1.50 & 1.04 & 4.81 & 3.23 & 2.58 & 1.66 \\
MEGAN       & 2.03 & 1.76 & 1.50 & 1.01 & 4.82 & 3.26 & 2.59 & 1.67 \\
Chemformer  & 1.77 & 1.02 & 0.68 & 0.35 & 4.70 & 1.73 & 1.05 & 0.53 \\
Graph2Edits & 2.11 & 1.81 & 1.54 & 1.09 & 4.80 & 3.12 & 2.40 & 1.51 \\
MHNreact    & 2.08 & 1.86 & 1.63 & 1.28 & 4.84 & 3.34 & 2.75 & 1.98 \\
RetroKNN    & 2.13 & 1.86 & 1.62 & 1.23 & 4.84 & 3.37 & 2.75 & 1.90 \\
R-SMILES    & 2.10 & 1.87 & 1.64 & 1.24 & 4.87 & 3.51 & 2.86 & 1.95 \\
\midrule
\multicolumn{9}{c}{\textit{\ourmodel{}}, Supervised Fine-Tuning, \textbf{Top-1 Mode}} \\
\midrule
\ourmodel{}-LFM2-CREED-CCV+USPTO\textsuperscript{*}              & 1.63 & 1.08 & 0.74 & 0.39 & 4.15 & 2.17 & 1.42 & 0.74 \\
\midrule
\multicolumn{9}{c}{\textit{\ourmodel{}}, Supervised Fine-Tuning, \textbf{Top-K Mode}} \\
\midrule
\ourmodel{}-LFM2-CREED-CCV+USPTO         & 2.01 & 1.72 & 1.45 & 1.00 & 4.23 & 2.84 & 2.28 & 1.51 \\
\ourmodel{}-LFM2-CREED-CCV-2+USPTO-XL    & 2.07 & 1.82 & 1.60 & 1.27 & 4.17 & 2.92 & 2.43 & 1.78 \\ 
\midrule
\multicolumn{9}{c}{\textit{\ourmodel{}}, Reinforcement Learning Fine-Tuning, \textbf{Top-K Mode}} \\
\midrule
\ourmodel{}-LFM2-RFT-CC          & 2.08 & 1.86 & 1.65 & 1.29 & 4.16 & 2.96 & 2.47 & 1.79 \\
\ourmodel{}-LFM2-RFT-CC-NR       & 2.18 & 1.95 & 1.74 & 1.38 & 4.29 & 3.05 & 2.56 & 1.92 \\
\bottomrule
\end{tabular}

}
\caption{Plausibility-based evaluation in the single-step retrosynthesis Top-$K$ mode. \textbf{Max}: per-target maximum ChemCensor score averaged over TMs. \textbf{Av.\ PT-Top-K CC}: per-TM average ChemCensor score over top-K unique predictions; \textbf{ChemCensor v0.5.2.}}
\label{tab:main_table_topk_052}
\end{table*}
\clearpage
\section{Distribution of Reactant Sets per Product in USPTO-50K-test-mini}
\label{app:dist_average}

\begin{figure}[htbp]
  \centering
  \includegraphics[width=\columnwidth]{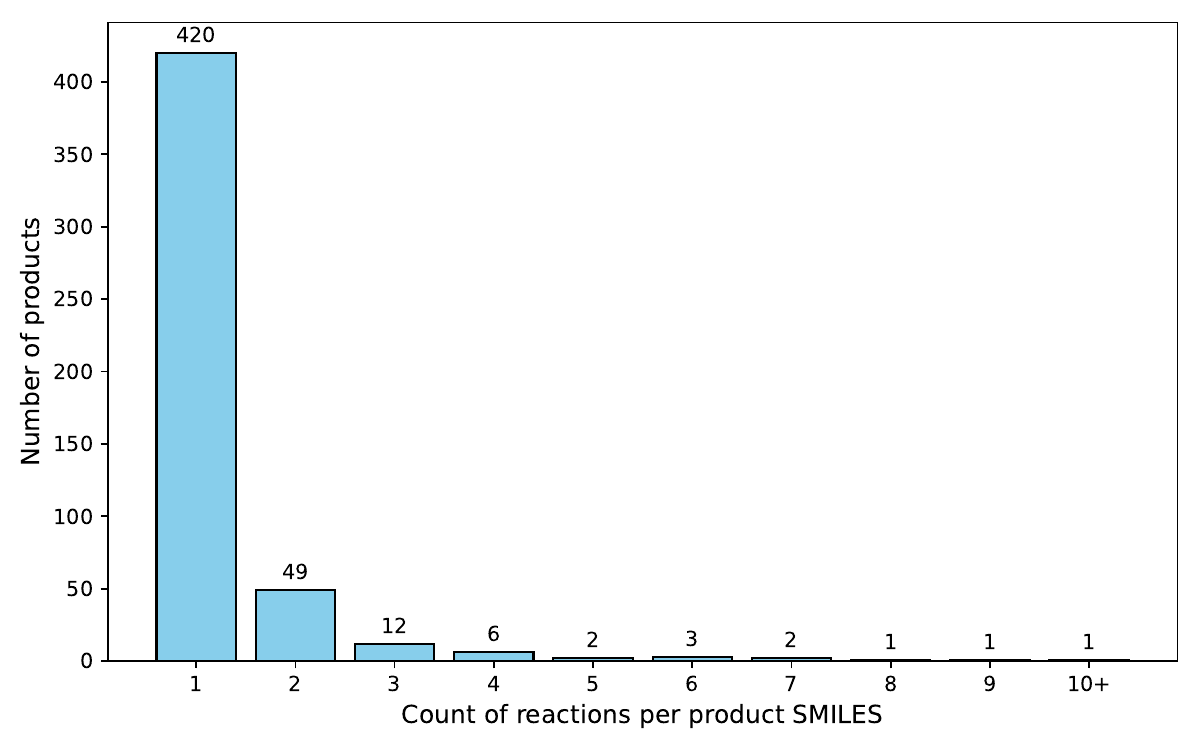}
  \caption{Distribution of the number of reactant sets for products from USPTO-50K-test-mini. For each product, all reference reactant sets were retrieved by matching it against USPTO-full.}
  \label{fig:reactant_dist}
\end{figure}

\autoref{fig:reactant_dist} shows the distribution of the number of distinct reactant sets (reference reactions) per product across the USPTO-50K-test-mini set. For each product, we collected all reference reactant sets by matching the product against the entire USPTO-full corpus \citep{Lowe2017usptofull}, rather than relying only on the single reaction provided in the test split. Even so, the large majority of products (420 of 497) are associated with a single reference reaction, and only a small tail has two or more. This sparsity of reference routes is the main limitation of exact-match, single-reference evaluation, and it motivates both our plausibility-based Top-K evaluation and the USPTO-XL augmentation, which enriches products with additional verified reactant sets.

\section{CREED-CCV-2+USPTO-XL Details}
\label{app:creed_ccv_2}

In this work, we introduce the CREED-CCV-2+USPTO-XL training dataset. Relative to the original CREED-CCV \citep{zagribelnyy2026chemcensor_c3lm}, it is substantially larger and denser: $3{,}680{,}906$ unique products and $45{,}649{,}785$ reaction candidates (vs. $698{,}765$ and $6{,}368{,}986$), i.e.\ roughly $5.3\times$ more products and $7.2\times$ more reactions, with $\sim$$12.4$ candidates per product on average (vs.\ $\sim$$9.11$). Partitions follow the same $0.8/0.1/0.1$ product-disjoint split. The per-product candidate count is concentrated in the $11$--$50$ bin ($2{,}099{,}509$ products), with the remainder in the $6$--$10$ ($785{,}011$), $2$--$5$ ($579{,}555$), $1$ ($216{,}754$), and $50+$ ($77$) bins.

\end{document}